\documentclass[10pt,twocolumn,letterpaper]{article}

\usepackage[pagenumbers]{cvpr}      %
\usepackage{multirow} 
\usepackage{bbding}
\usepackage{amssymb}   %
\usepackage{pifont}    %
\usepackage{tcolorbox}

\definecolor{cvprblue}{rgb}{0.21,0.49,0.74}
\usepackage{xcolor, soul}
\usepackage[pagebackref=true,breaklinks=true,colorlinks,bookmarks=false]{hyperref}
\hypersetup{linkcolor=[rgb]{0.7,0.1,0.1}}
\hypersetup{citecolor=[rgb]{0.4,0.15,0.95}}

\def\paperID{****} %
\def\confName{CVPR}
\def\confYear{2025}

\title{ChatDyn: Language-Driven Multi-Actor Dynamics Generation \\ in Street Scenes}

\author{
  Yuxi Wei$^1$ \quad Jingbo Wang$^2\footnotemark[2]$ \quad Yuwen Du$^1$ \quad Dingju Wang$^1$ \\ Liang Pan$^2$ \quad Chenxin Xu$^1$ \quad Yao Feng$^{3,4}$ \quad Bo Dai$^{2}$ \quad Siheng Chen$^1\footnotemark[2]$\\~
\small $^1$ Shanghai Jiao Tong University \quad
$^2$ Shanghai AI Laboratory \\
\small $^3$ Max Planck Institute for Intelligent Systems \quad $^4$ ETH Z\"urich\\
 \texttt{\footnotesize \{wyx3590236732, wangdingju, xcxwakaka, sihengc\}@sjtu.edu.cn}, \quad \texttt{\footnotesize duyuwen@tju.edu.cn} \\
 \texttt{\footnotesize \{wangjingbo, panliang, daibo\}@pjlab.org.cn}, 
 \quad \texttt{\footnotesize yao.feng@tuebingen.mpg.de}
 \\
 \href{https://vfishc.github.io/chatdyn/ }{\texttt{\small vfishc.github.io/chatdyn}}
}

\begin{document}
\maketitle
\renewcommand{\thefootnote}{\fnsymbol{footnote}}
\footnotetext[2]{Corresponding authors.}
\begin{abstract}
\quad Generating realistic and interactive dynamics of traffic participants according to specific instruction is critical for street scene simulation. However, there is currently a lack of a comprehensive method that generates realistic dynamics of different types of participants including vehicles and pedestrians, with different kinds of interactions between them. In this paper, we introduce ChatDyn, the first system capable of generating interactive, controllable and realistic  participant dynamics in street scenes based on language instructions. To achieve precise control through complex language, ChatDyn employs a multi-LLM-agent role-playing approach, which utilizes natural language inputs to plan the trajectories and behaviors for different traffic participants. To generate realistic fine-grained dynamics based on the planning, ChatDyn designs two novel executors: the PedExecutor, a unified multi-task executor that generates realistic pedestrian dynamics under different task plannings; and the VehExecutor, a physical transition-based policy that generates physically plausible vehicle dynamics. Extensive experiments show that ChatDyn can generate realistic driving scene dynamics with multiple vehicles and pedestrians, and significantly outperforms previous methods on subtasks. Code and model will be available at \href{https://vfishc.github.io/chatdyn/ }{\texttt{\small https://vfishc.github.io/chatdyn}}.

\end{abstract}
    
\section{Introduction}
\label{sec:intro}

\quad 
Autonomous driving simulation plays a critical role in the training and validation of driving systems, which attracts significant attention in recent years. To create a realistic simulation system, it must replicate real world scenarios, including the dynamics of various traffic participants such as pedestrians and vehicles~\cite{galvao2024pedestrian, golchoubian2023pedestrian, gulzar2021survey}, as well as their interactions.  These dynamics are essential for constructing realistic scene events, maintaining the temporal consistency of simulations~\cite{caesar2020nuscenes, sun2020scalability, xiao2021pandaset, dosovitskiy2017carla}, and enabling accurate decision making~\cite{amado2020pedestrian, ni2016evaluation, thakur2019assessment} in autonomous systems. 
However, generating such dynamics is challenging, as it involves modeling the behavior of vehicles and pedestrians individually while also capturing their intricate interactions within street scenes. This requires high-level planning to coordinate trajectory generation for both types of participants, taking the street scenes into account, and precise low-level control to ensure that their interactions appear realistic. 

\begin{figure}[t]
    \centering
    \vspace{-1mm}
\includegraphics[width=0.44\textwidth]{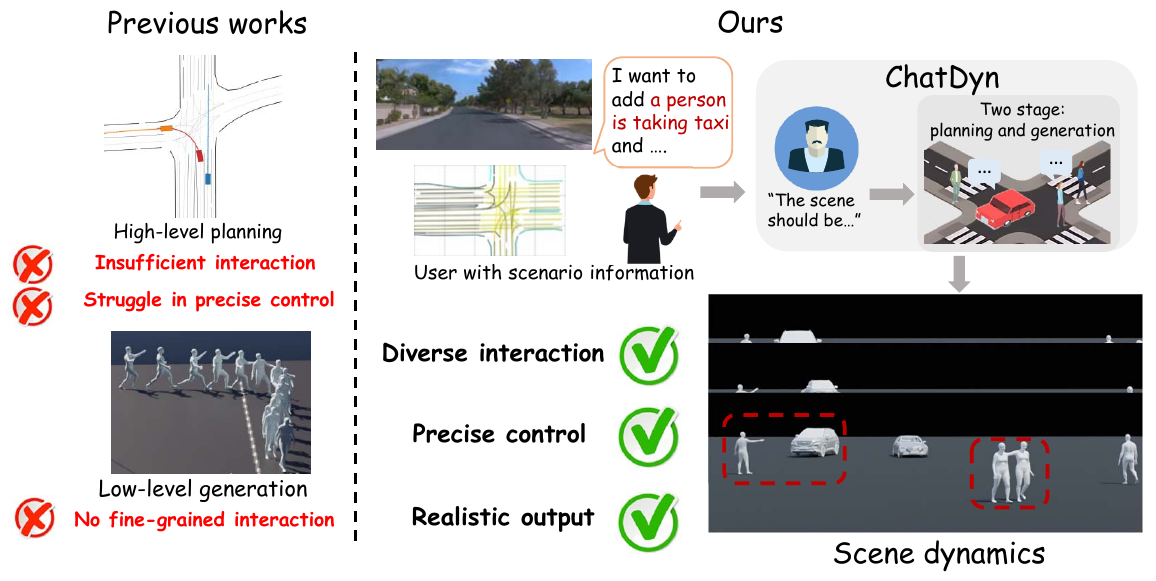}
    \vspace{-3mm}
    \caption{\small ChatDyn achieves interactive and realistic language-driven multi-actor dynamics generation in street scenes.
    }
    \label{fig:teaser}
    \vspace{-4mm}
\end{figure}

Existing methods often focus on one of these aspects while neglecting the other. High-level planning approaches, such as LCTGen~\cite{tan2023language} and CTG++\cite{zhong2023language}, use language-based inputs to generate scene-level plans. However, they struggle to control detailed participant behaviors and often lack diversity in participants, such as pedestrians, as well as the varied interactions between different participants. On the other hand, low-level generation methods, such as Pacer\cite{rempe2023trace} and Pacer+~\cite{wang2024pacer+}, excel at producing fine-grained pedestrian dynamics based on specific control signals. Yet, they fail to account for the nuanced interactive behaviors that arise between pedestrians and other participants.

In this work, we introduce ChatDyn, the first system achieving interactive and realistic language-driven multi-actor dynamics generation in street scenes, as shown in Fig.~\ref{fig:teaser}. 
The key insight behind ChatDyn is the need to combine both high-level planning and low-level control to create a realistic simulation system. Large language models (LLMs) excel at high-level planning, while physics-based methods are well-suited for fine-grained, low-level control. Building on this observation, ChatDyn combines a multi-LLM agent for high-level planning with physics-based executors for low-level generation. 
By integrating high- and low-level components, ChatDyn allows users to input language instructions to specify their requirements. It then generates realistic dynamics for diverse traffic participants, including pedestrians and vehicles, within the specified scene, while accommodating complex and varied interactions among participants. 

Specifically, ChatDyn employs multi-LLM-agent role-playing for high-level planning to understand complex and abstract user instructions. 
The key idea is to treat each traffic participant as an agent, which can extract specific requirements and semantics from the language instructions, interact according to the requirement of scenes, and generate planned trajectories and behaviors with interactive characteristics. The multi-LLM-agent role-playing approach shows two notable advantages: first, LLM has powerful language understanding capabilities, enabling them to accurately capture the complex and abstract demands in user instructions, thus producing planning results that closely align with them; second, the interaction between agents allows for the incorporation of interactive information during the planning process, ensuring that the planning results, at the high-level stage, exhibit interactive characteristics. 

Due to the different characteristics of the dynamics of pedestrians and vehicles, we design separate executors for their low-level generation.
To generate realistic and interactive fine-grained low-level pedestrian dynamics, we propose PedExecutor, a unified method accomplishes multiple high-level planned tasks, including fine-grained physical interactions, and returns human-like dynamics. PedExecutor is a physics-based control policy that fulfills multiple control requirements and enables interactive tasks for pedestrians under realistic physical feedback. A task masking mechanism and related training strategy make it handle different requirements with a single unified policy. PedExecutor also incorporates hierarchical control and body-masked adversarial motion prior, which introduce priors to both the action space and reward space, allowing for the generation of realistic, natural dynamics while executing the planning.

To obtain physically feasible low-level vehicle dynamics, we propose VehExecutor, a control policy based on vehicle physical transition process. VehExecutor generates vehicle actions from high-level trajectory planning, and physical transition process returns the next vehicle state which can be accumulated as final dynamics output. VehExecutor ensures the physical feasibility of output dynamics and eliminating unrealistic factors such as tail swings and unwanted translations. Additionally, VehExecutor incorporates history-aware state and action design, simulating the real-world decision-making process of vehicles, which improves the accuracy and consistency of control.

Based on the above designs, ChatDyn satisfied the three required characteristics: for interactivity, multi-LLM-agent role-playing enables interaction planning at the high-level, while the executors execute the interaction-aware plans to generate low-level interactive behaviors; for controllability, LLM-agents precisely follow the user's natural language inputs, and the executors precisely execute the planning for accurate control; for realism, PedExecutor employs hierarchical control and integrates body-masked adversarial motion priors, producing dynamics that align with human intuitions, while VeExecutor highly simulates the physical feedback and kinematic characteristics of real vehicle.

Our contributions can be summarized as: (1) We design the first language-driven dynamics generation system for multi-actor in street scenes, utilizing multi-LLM-agent role-playing for high-level planning. (2) We propose a unified pedestrian executor that, under hierarchical control, executes multiple planning tasks and generates realistic low-level pedestrian dynamics based on planning. (3) We propose a vehicle executor that derives realistic control processes and low-level vehicle dynamics from planning trajectories. (4) We demonstrate the realistic language-driven outputs of the system in experiments and validated the effectiveness of its each individual component.

\section{Related Works}
\label{sec:related_works}
\quad
\textbf{Human dynamics generation.}
The generation of human dynamics can broadly be divided into kinematics-based and physics-based approaches. For kinematics-based generation, transformer-based methods~\cite{athanasiou2022teach, guo2024momask, guo2022generating} and diffusion models~\cite{zhang2022motiondiffuse, zhang2023remodiffuse, dabral2023mofusion, kim2023flame} are used to generate corresponding dynamics based on language inputs. Recent methods \cite{shafir2023human, xie2023omnicontrol, wan2023tlcontrol} further introduce more control conditions,  while others~\cite{zhou2025emdm, dai2024motionlcm} focus on improving efficiency in the generation process. However, these approaches do not account for physical constraints, often resulting in physically implausible results, particularly for interactions.
For physics-based generation, existing work like \cite{peng2021amp, peng2022ase, dou2023c, won2022physics, yao2022controlvae, luo2023perpetual, luo2023universal} achieve predefined tasks with plausible dynamics. \cite{tessler2023calm, juravsky2022padl, bae2023pmp, xu2023composite, xu2023adaptnet} extend the human character dynamics content with different designs. \cite{rempe2023trace, wang2024pacer+} focus on the dynamics of street scene pedestrians. However, these methods do not involve the interaction behaviors between multi-pedestrians. Our PedExecutor considers the interaction behaviors, and is trained as a unified policy for multiple scenarios also including following, imitation.

\textbf{Vehicle dynamics generation.}
In industry, vehicle dynamics are usually generated by software tools~\cite{chen2022generating, queiroz2019geoscenario, fremont2019scenic, jesenski2019generation}. Some research works~\cite{bergamini2021simnet, tan2021scenegen, feng2023trafficgen, rempe2022generating} focus on unconditioned generation to simplify the generation process.
Recent works on vehicle dynamics generation under different conditions have introduced various approaches. TrafficSim~\cite{suo2021trafficsim} uses predictive models to generate vehicle dynamics in a scene, while CTG~\cite{zhong2023guided} introduces diffusion models into the generation process. CTG++~\cite{zhong2023language} and LCTGen~\cite{tan2023language} achieve scene-vehicle dynamics generation controlled by language, and SceneControl~\cite{lu2024scenecontrol} adds multiple controllable parameters to the generation process. RealGen~\cite{ding2023realgen} uses a retrieval-based approach to improve the realism of generated dynamics. However, these works generally do not directly consider physical constraints and kinematic constraints, and lack sufficient interaction. And these language-controlled methods cannot achieve precise control over different vehicles. Our LLM-agents planning utilizes the language understanding capability of LLMs to achieve precise control of participants considering interactive information, and VehExecutor generates the final physical plausible vehicle dynamics.

\textbf{Large language models and agents.}
Recently, numerous large language models~\cite{touvron2023llama, liu2024visual, bai2023qwen, achiam2023gpt, yang2023baichuan, glm2024chatglm} are proposed and released. Many works have leveraged these models by finetuning them\cite{hu2021lora, liu2024boft, Qiu2023OFT, feng2024chatpose} or integrating them with relevant tools to build LLM-based agents~\cite{liu2023agentbench, wu2023autogen} capable of performing targeted tasks or specific functions.  These agents have been applied to a wide range of downstream tasks, including code development~\cite{hong2023metagpt, hou2023large}, web browsing~\cite{zhou2023webarena, koh2024visualwebarena}, medical Q\&A~\cite{li2024agent, li2024mmedagent}, social behavior simulation~\cite{pang2024self, leng2023llm}, and issue handing~\cite{shen2024hugginggpt, tao2024magis}. In this paper, we explore the application of agents in traffic simulation, utilizing them to interact and perform trajectory and behavior planning within traffic scenarios.

\section{Method}

\begin{figure}[t]
    \centering
    \vspace{-2mm}
\includegraphics[width=0.42\textwidth]{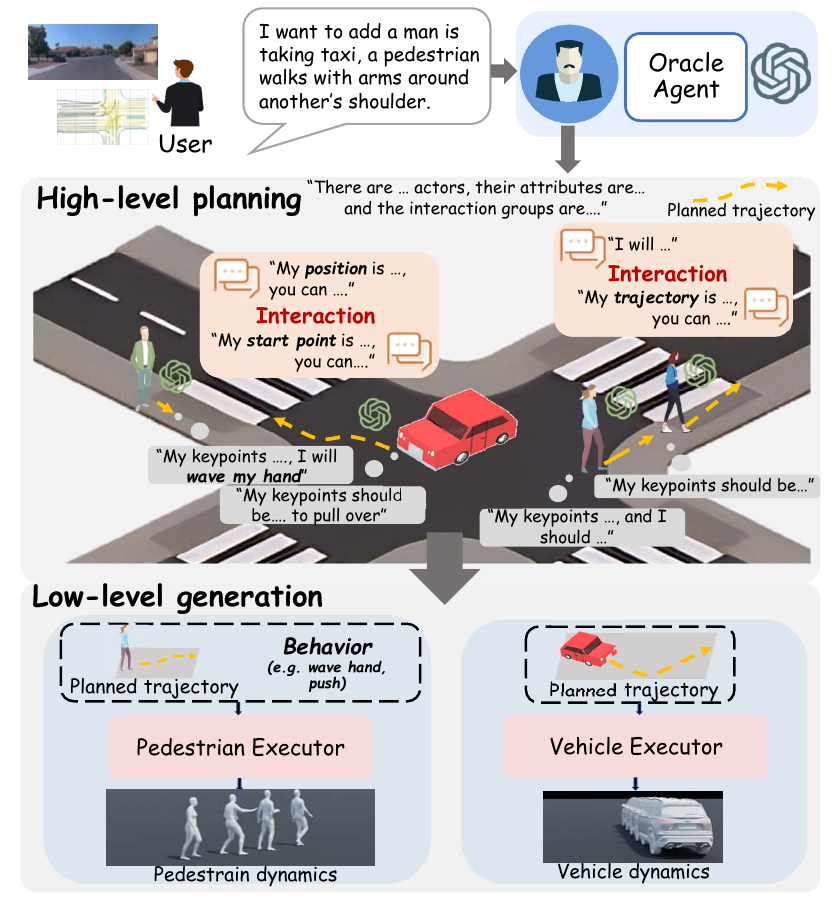}
    \vspace{-3mm}
    \caption{\small System overview. ChatDyn adopts multi-LLM-agent role-playing for precise high-level planning. Two specialized executors are designed for realistic low-level generation.
    }
    \label{fig:overview}
    \vspace{-3mm}
\end{figure}
\subsection{System Overview}
\quad
ChatDyn interprets and analyzes user language instructions, then produces scene dynamics that align with them. ChatDyn employs a two-stage process: high-level planning which plans trajectory and behavior under complex and abstract command; low-level generation for fine-grained, realistic dynamics generation. Since user instructions may contain many specific details that require precise control and abstract semantics that need to be understood, ChatDyn leverages multi-LLM-agent role-playing, treating each traffic participant as an LLM-agent. This approach capitalizes on the LLM's ability to comprehend semantic information and its extensive commonsense priors, using specific tools 
and interaction process to complete high-level trajectory and behavior planning. Each traffic participant's corresponding agent is also equipped with an executor as one of the tools. After the high-level planning is completed, the executor uses the planning results to execute the low-level generation process. The executors generate fine-grained, realistic, and physically feasible dynamics based on high-level planning. See Fig. \ref{fig:overview} for illustration.

\subsection{High-Level Planning} 
\quad
High-level planning is completed by LLM-agents and interaction among them. Each LLM-agent consists of an LLM component and the relevant tools, and the agents are divided into two types: oracle agents and actor agents. The oracle agent is responsible for directly receiving user instruction and understanding, dispatching them. Oracle agent breaks the instructions down into specific information for each participant, identifies interaction relationships, forms interaction groups, and generates the planning order schedule for the following process. Each actor agent corresponds to a specific traffic participant, initialized with the information provided by the oracle agent. The actor agents then interact with others based on the interaction groups to get information from others, and finally plan trajectory and behavior using their tools. The following parts will provide a detailed description of these two types of agents.

\subsubsection{Oracle Agent}
\quad
Oracle agent breaks down complex and lengthy user instructions into specific actions for each participant, ensuring an accurate understanding of the user’s instructions and generating the subsequent schedule. For its LLM component, we explain its responsibilities, outline our requirements and give some few-shot examples. The LLM component of the oracle agent outputs the initialization information for each participants, interaction groups, and the schedule in natural language. To facilitate subsequent execution, we equip it with a structured output tool that converts language output into a predefined data structure for execution. Oracle agent provides the system with the ability to handle complex, mixed, and abstract instructions, and streamlines operations for clarity and fine granularity. Detailed instruction prompt can be found in the appendix.

\begin{figure*}[t]
    \centering
    \vspace{-3mm}
\includegraphics[width=0.9\textwidth]{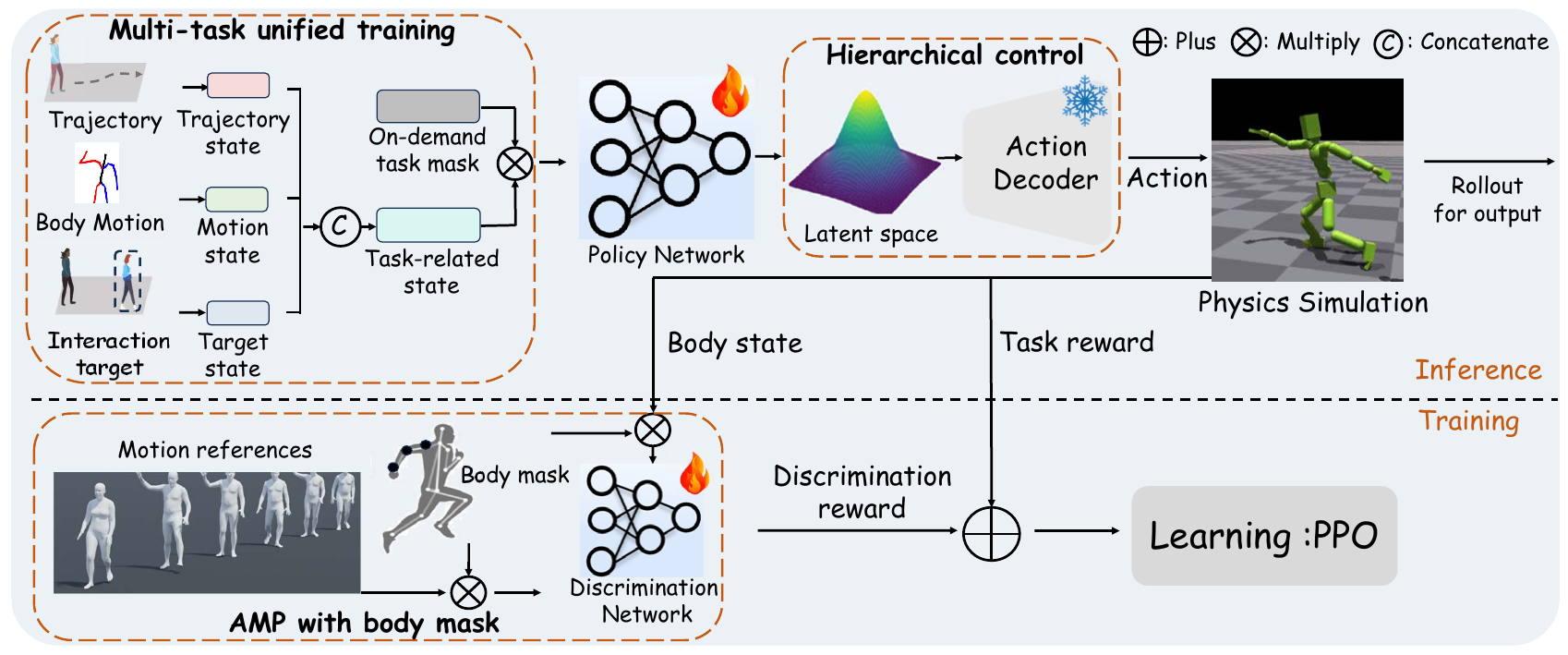}
    \vspace{-2mm}
    \caption{\small Pedestrian executor (PedExecutor) framework. With multi-task unified training, PedExecutor achieves unified control over various task, including following, imitation and interaction. It generates realistic pedestrian dynamics by effectively executing tasks derived from planning. Hierarchical control and AMP with body mask provide prior to action space and reward space, improving the realism of dynamics output.}
    \vspace{-3mm}
    \label{fig:pedestrian_method}
\end{figure*}

\subsubsection{Actor Agent} 
\quad
Each actor agent corresponds to a specific participant in the scene and is initialized with information provided by the oracle agent, including the agent type and description about its trajectory and behavior. The actor agents interact based on the oracle agent's schedule to collect interactive information and completes the final high-level planning. The planning process of the actor agent incorporates map information, with the scene map represented as a graph \(\mathcal{G} = (\mathcal{N}, \mathcal{E})\), where \(\mathcal{N}\) represents different lane sections. Each node in \(\mathcal{N}\) contains the point set for the lane section and other extracted information such as its orientation and driving type (e.g., straight, left/right turn, lane/boundary). The edges \(\mathcal{E}\) represent the relationships between the lane sections, such as adjacency or connectivity. The specific planning process uses a keypoints-based approach, integrating actor interactions to complete the planning.

\textbf{Keypoints-based trajectory planning.} Intuitively, a trajectory is typically determined by a series of keypoints, which are then interpolated to form the complete trajectory. Specifically, the actor agent first obtains keypoints and then uses interpolation to get the planned trajectory. The LLM component of the actor agent infers the number of keypoints needed based on its initialization information, and then determines the generation process for each keypoint. There are generally two sources for keypoints: one is directly obtained from map  and previous keypoints, where the actor agent retrieves keypoints from the map \(\mathcal{G}\) based on the keypoint's attributes and relationship with the previous keypoint. The other source is through interaction, where a keypoint depends on information from others. This kind of keypoints are determined by actor interaction.

\textbf{Multi-actor interaction. }  Interactions allow agents to acquire information from other agents to generate certain keypoints, thereby fulfilling the interaction requirements during the high-level planning process. The groups that need to interact have already been specified by the oracle agent. During the interaction, the LLM determines which pieces of information are needed from the interacting agents for a specific keypoint, such as the start point, endpoint, or the entire trajectory. Each agent then shares the required information, and this exchanged information is provided to the corresponding tools that generate keypoints based on the agent's own needs. Actor interaction enables interactive trajectory generation, which is common in the real scenarios.

After obtaining all desired keypoints, interpolation is applied to generate a complete trajectory, fulfilling the trajectory planning stage. Here, we employ Bézier curves for interpolation. Note that: (i) the interpolated trajectory at this stage lacks physical constraints and serves only as a preliminary plan to be executed by the subsequent executor; (ii) for static actors, two keypoints are required to determine orientation, as their facing direction must be specified; (iii) when the actor type is pedestrian, besides trajectory, additional behavioral instructions may be inferred by LLM component based on the initialization information. These behaviors could stem from direct command-based controls or indirectly from semantic cues, with the final instructions provided in text form for the executor to execute.

\subsection{Low-Level Generation for Pedestrian}
\quad
The pedestrian executor (PedExecutor), as shown in Fig. \ref{fig:pedestrian_method}, generates low-level pedestrian dynamics based on the trajectories and behaviors planned from high-level planning. Pedestrian behaviors can be subdivided into single-agent behavior directly specified by language and interactive behaviors that occur between multiple agents. Thus the challenge lies in simultaneously handling trajectory following, single-agent motion specification, and multi-agent interactions, while maintaining human-like quality. To achieve these, PedExecutor utilizes multi-task unified training to execute the trajectory, single-agent behavior, and multi-agent interactions as planned by the LLM with a single policy. For human-like quality, the action space incorporates hierarchical control to provide priors, while the reward function uses body masked Adversarial Motion Priors (AMP)~\cite{peng2021amp} to encourage human-like control. Ultimately PedExecutor returns realistic dynamics that follow planned trajectory and complete desired behaviors.

Control process is defined by Markov decision process \( \mathcal{M}^{p} = \{\mathcal{S}^{p}, \mathcal{A}^{p}, \mathcal{T}^{p}, \mathcal{R}^{p}, \gamma^{p}\} \), where the elements represent states, actions, transition process, reward, and the discount factor. Goal-conditioned reinforcement learning (RL) is adopted for training. The transition process is implemented by the physics engine. Others are detailed below.

\textbf{States and multi-task unified training. }
Our control policy is designed to handle multiple distinct tasks, requiring specialized processing and training to achieve a unified policy capable of addressing different tasks. The tasks are divided into three main categories based on requirements: trajectory following, single-agent behavior specification, and multi-agent interaction. For single-agent behavior, the LLM generates corresponding text descriptions, which are then processed by a Text2Motion model (e.g., MoMask~\cite{guo2024momask}) to produce an upper-body motion for the policy to imitate thereby achieve single-agent behavior specification. Multi-agent interaction tasks are trained according to predefined types, with the LLM's planning classifying interaction behaviors into these types, enabling the executor to execute specific interactions as specified.

Task-related states consist of trajectory slices to be followed \( \mathcal{S}^{p}_{traj} \), motion to be imitated \( \mathcal{S}^{p}_{mo} \), and states of interacting targets \( \mathcal{S}^{p}_{tar} \). \(\mathcal{S}^{p}_{traj} \) includes K future steps of planned trajectory from the current timestep, while \( \mathcal{S}^{p}_{mo} = \text{concat}(\hat{j}_{pos}, \hat{j}_{rot}, \hat{j}_{vel}, \hat{j}_{\omega}) \) comprises joint position \( \hat{j}_{pos} \), joint rotation \( \hat{j}_{rot} \), joint velocity \( \hat{j}_{vel} \), and rotation velocity \( \hat{j}_{\omega} \) from the motion, with optional joint masking to retain only relevant parts, such as upper body. \( \mathcal{S}^{p}_{tar} = \text{concat}(r_{pos}, r_{rot}, r_{vel}, r_{\omega}, r_{bbox}, e_{inter}) \) includes the interacting target’s root position \( r_{pos} \), root rotation \( r_{rot} \), root velocity \( r_{vel} \), root angular velocity \( r_{\omega} \), and the bounding box’s 8 vertices \( r_{bbox} \). Additionally,  an interaction task embedding \( e_{inter} \) represented as a one-hot vector to specify the interaction type. 
Beyond task-related states, the final observations also include humanoid proprioception~\cite{wang2024pacer+} \( \mathcal{S}^{p}_{prop} \) to capture necessary internal states of the humanoid. The final $\mathcal{S}^{p} = \text{concat}(\mathcal{S}^{p}_{traj}, \mathcal{S}^{p}_{mo}, \mathcal{S}^{p}_{tar}, \mathcal{S}^{p}_{prop})$

To activate specific parts during training/testing, we introduce a task masking mechanism. During training, tasks to be executed for each episode are sampled, and the remaining tasks related states are masked. Only the relevant environment for the involved tasks is prepared, and rewards are calculated based on that. During inference, states are activated as needed, while others are masked, enabling task-specific execution.
This approach ensures the trained unified policy handles all of the tasks without confusion. After sufficient training, its performance matches that of policies trained on individual tasks, while can also execute multiple non-conflicting tasks in a single run.

\textbf{Action hierarchical control.}
Pedestrian action spaces are typically modeled using a proportional-derivative (PD) controller at each degree of freedom (DoF), but such spaces lack inherent priors, often leading to locally unrealistic actions for completing specific tasks. To address this, we apply a hierarchical action control space from PULSE~\cite{luo2023universal}. The policy network first outputs to a pretrained latent space, obtaining a latent feature \(\mathbf{f_{action}}\), which is then decoded by a pretrained decoder into control signals. The pretrained latent space, equipped with the corresponding decoder, ensures that the decoded output distribution closely matches the input data from pretraining, thereby providing the action space with a real-world action prior.

\textbf{Reward design. }
The reward consists of two main components. The first is the discrimination reward $R_{disc}$ used to implement AMP~\cite{peng2021amp}, which employs a discriminator to encourage the policy to generate output that align with movement patterns observed in a dataset of human-recorded data clips. The second component is the task-related reward $R_{task}$ , designed to motivate the policy to accomplish specific tasks for executing high-level planning. The detailed task reward designs can be found in appendix. Similar to \cite{wang2024pacer+}, we implement early termination if excessive root distance error or joint distance error occurs during following or imitation. 
\begin{figure}[t]
    \centering
    \vspace{-2mm}
\includegraphics[width=0.45\textwidth]{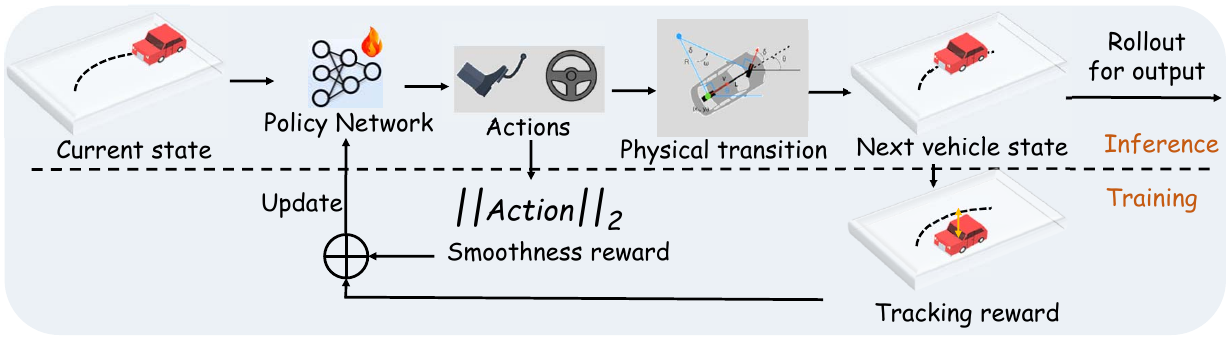}
\vspace{-1mm}
    \caption{\small Vehicle executor (VehExecutor) framework. VehExecutor adopts goal-conditioned RL based on physical transition of real vehicle. Combining with history-aware design, VehExecutor generates realistic vehicle dynamics under planned trajectory.
    }
    
    \label{fig:vehicle_method}
    \vspace{-4mm}
\end{figure}

\begin{figure*}[t]
    \centering
    \begin{minipage}{0.45\textwidth} %
        \centering
        \vspace{-3mm}
        \includegraphics[width=\textwidth]{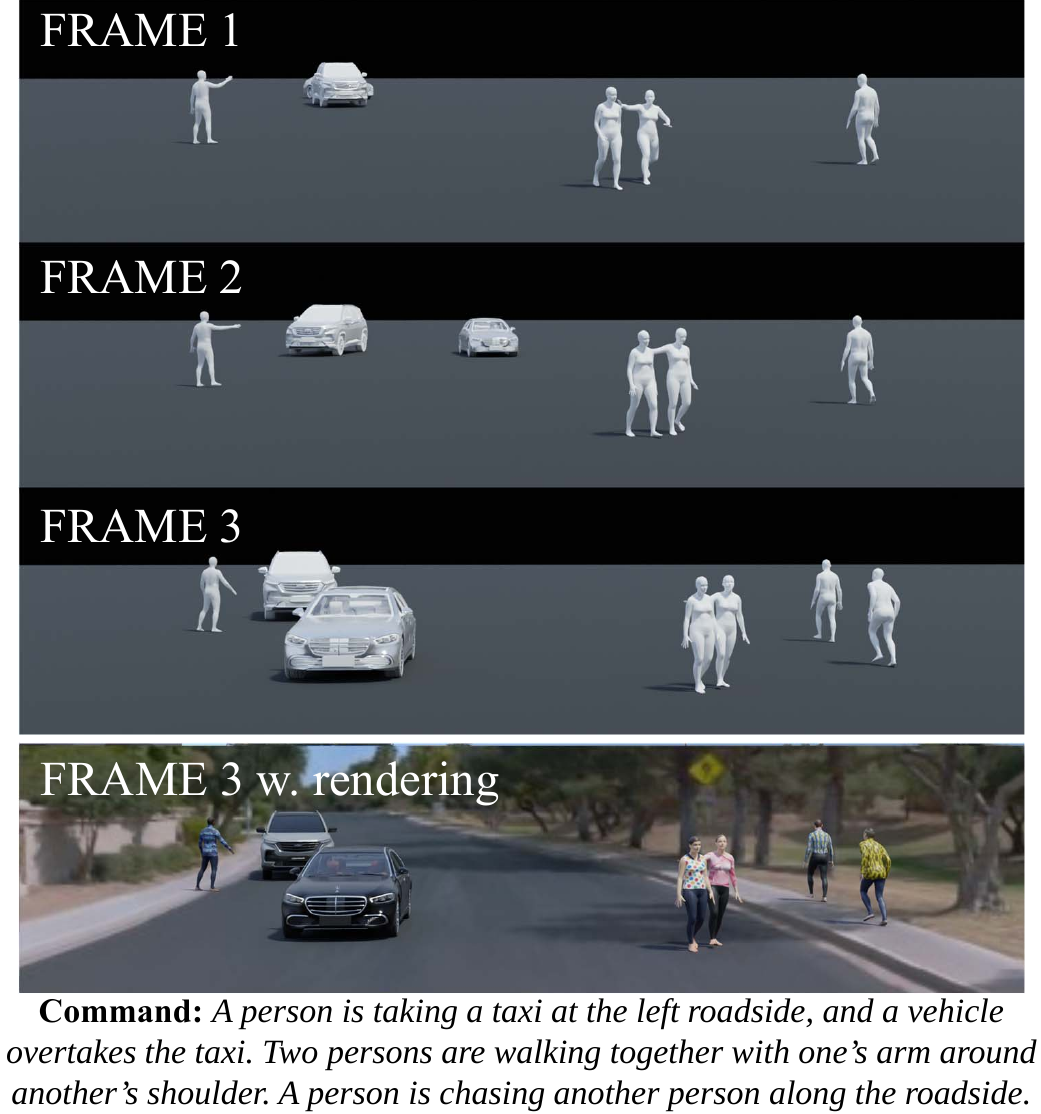} %
    \end{minipage}
    \hfill
    \begin{minipage}{0.45\textwidth} %
        \centering
        \vspace{-3mm}
        \includegraphics[width=\textwidth]{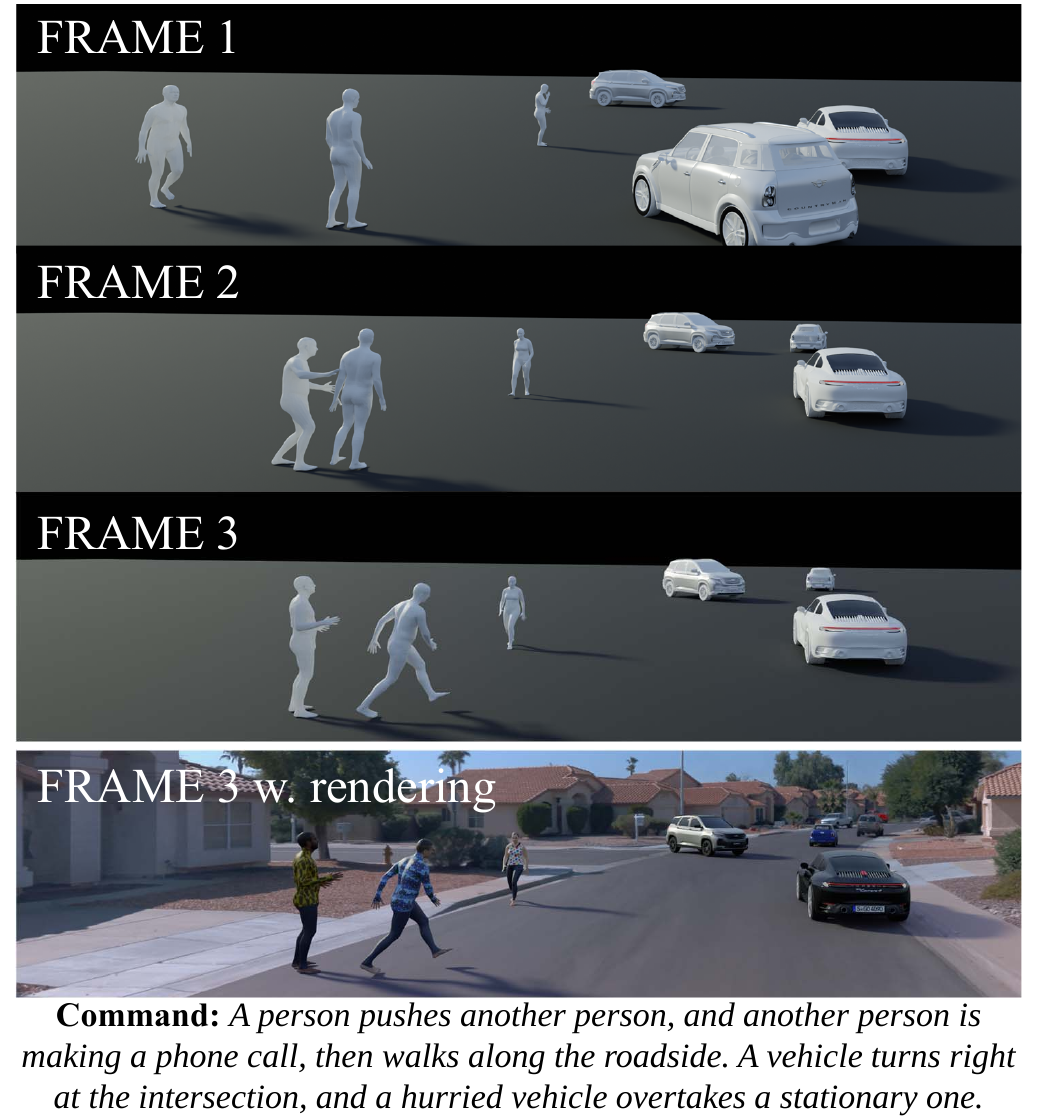} %
    \end{minipage}
    \vspace{-1.5mm}
    \caption{System results under complex and composite commands, with diverse interaction information and realistic dynamics output.}
    \vspace{-5mm}
    \label{fig:system_result}
\end{figure*}
\textbf{AMP with body mask and warm-up training. }
We employ Proximal Policy Optimization (PPO)~\cite{schulman2017proximal} for overall training optimization. However, directly optimizing for interaction tasks presents certain challenges: without using AMP, the lack of a discrimination reward makes it difficult to achieve human-like dynamics results. Conversely, proper reference data clip needed for specific tasks are usually complex and not easily obtainable. Using reference data clip from the following or imitation tasks can lead to misalignment with the actual requirements for interaction, resulting in a policy that fails to accomplish interaction tasks.

To address these issues, we adopt two training techniques: (i) Initially, we train without AMP. While this approach may not yield realistic dynamics, it allows the policy to complete interaction tasks, providing a more accurate initial domain for subsequent optimization. Once this foundation is established, we incorporate AMP training, which maintains task completion while improving dynamics quality. (ii) We introduce AMP body mask, where certain joints are excluded during the AMP calculations. Since interaction tasks primarily engage specific body part(e.g. arm), we apply the AMP mask to the relevant body parts, ensuring that the rest parts of dynamics can leverage AMP for more natural movements while allowing the specific part to complete the tasks.
By utilizing these two training techniques, we achieve a policy that satisfies both visual realism and the ability to perform interaction tasks effectively.

\subsection{Low-Level Generation for Vehicles}

\quad The vehicle executor (VehExecutor), as shown in Fig. \ref{fig:vehicle_method}, generates the final realistic and physically feasible vehicle dynamics with control policy based on the high-level planned trajectory, which may initially violate certain dynamic constraints. To involve physical constraints and achieve precise control, VehExecutor utilizes physics-based transition environment, combined with history-aware state and action space design. The final dynamics can be obtained by accumulating the vehicle position and heading from environment. The control process also modeled as a Markov decision process defined by \( \mathcal{M}^{v} = \{\mathcal{S}^{v}, \mathcal{A}^{v}, \mathcal{T}^{v}, \mathcal{R}^{v}, \gamma^{v}\} \), with goal-conditioned RL for training. Further details are provided below.

\textbf{History-aware states. }
The VehExecutor incorporates historical information into its states, alongside its current state, to improve the temporal consistency of the policy. The vehicle states consist of the planned trajectory segment \( \mathbf{\hat{P}}^{v} \), temporal velocity \( \mathbf{V}^{v} \), and dynamic parameters \( \Theta^{v} \). \( \mathbf{P}^{v} \) is a slice of the planned trajectory in the vehicle's coordinate system over the current and adjacent \( \tau^{v} \) frames. \( \mathbf{V}^{v} \) represents the vehicle's centroid velocity over \( \tau^{v} \) frames. \( \Theta^{v} \) includes inherent vehicle parameters such as \( L \) (vehicle length), \( W \) (vehicle width), \( l_f \) (front overhang) and \( l_r \) (rear overhang). These parameters provide prior information that influence the physical transition process.

\textbf{Vehicle actions. } 
To accurately simulate real vehicle actions and maintain temporal consistency, the vehicle's action space \( \mathcal{A}^{v} \) is defined by the delta steering angle \( \Delta \delta \) and scalar acceleration \( a \), which are the two most direct controls affecting vehicle movement in actual driving.

\textbf{Vehicle transition function.} 
We employed the bicycle model for modeling the vehicle physical transition process. Let \(x\) and \(y\) denote the vehicle's coordinates, \(v\) represent the scalar velocity, and \(\phi\) indicate the vehicle's orientation. Utilizing the inherent parameters from the states, the vehicle's state transition process can be expressed as: 
\begin{equation}
\begin{aligned}
    &\dot{x} = v*cos(\phi + \beta), \quad\dot{y} = v*sin(\phi + \beta), \\
    &\beta = arctan(\frac{l_r}{l_f+l_r}tan(\delta)), \quad \dot{\phi} = \frac{v}{l_f+l_r}cos(\beta)tan(\delta).
\end{aligned}
\nonumber
\end{equation}
Where $\beta$ denotes tire slip angle and $\dot{}$ denotes derivation. The bicycle model effectively simulates the physical state changes of the vehicle as determined by vehicle actions.

\textbf{Reward and training.}
The reward focuses on two key aspects: following the planned trajectory and the smoothness of results. Therefore, the reward consists of two components: \( R^{v}_{pos} = -||\hat{p}_{t} - p_{t}||_{2} \) for following the planned trajectory and \( R^{v}_{act} = -(||\Delta \delta||_{2} + ||a||_{2}) \) for smooth driving. Training uses TD3~\cite{fujimoto2018addressing} to maximize the accumulated discounted reward. Note that: (i) actions can be further smoothed with temporal filtering: \( \mathcal{A}^{v}_{t} = \alpha \mathcal{A}^{v}_{t-1} + (1 - \alpha) \mathcal{A}^{v}_{policy} \), where \( \mathcal{A}_{t} \) is the action taken at timestep \( t \) and \( \mathcal{A}_{policy} \) is the action directly output from the policy network; (ii) obstacles can be considered by concatenating their positions and radius in the state and adding \( R^{v}_{obs} = \frac{\epsilon}{L_o} \) to the reward if \( L_o < D \), where \( L_o \) is the distance to the obstacle, \(\epsilon\) is a coefficient and \( D \) is a threshold.

\section{Experiments}
\begin{figure}[t]
    \centering
\includegraphics[width=0.47\textwidth]{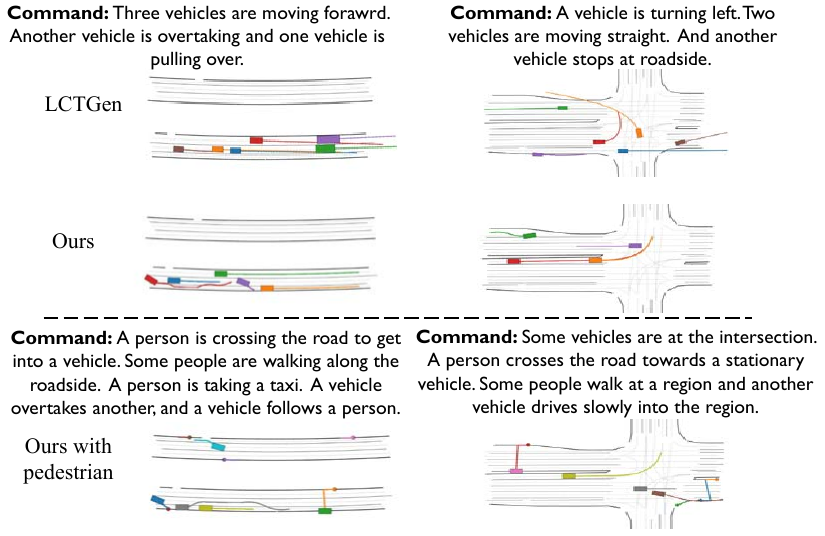}
\vspace{-3mm}
    \caption{\small High-level planning results comparison under vehicle-only command, and our planning results with pedestrians involved. The boxes indicate vehicles and the circles indicate pedestrians.
    }
    \label{fig:planning_compare}
    \vspace{-1mm}
\end{figure}

\begin{table}[t]
\centering
\footnotesize

\resizebox{.99\columnwidth}{!}{
\begin{tabular}{c|ccccc}
\toprule
\multirow{2}*{Methods}  &   \multicolumn{3}{c}{Language command category} & \multirow{2}*{{Within road}} & \multirow{2}*{{User preference}}\\
~ & single & interaction & compound & ~  & ~ \\
\midrule
LCTGen~\cite{tan2023language} & 0.926 & 0.209 & 0.682 & 0.604 & 0.143 \\
ChatSim~\cite{wei2024editable} & 0.874 & 0.052 & 0.778 & 0.865 & 0.061 \\
Ours & \textbf{0.952} & \textbf{0.883} & \textbf{0.896} & \textbf{0.935} & \textbf{0.796} \\
\bottomrule
\end{tabular}
}
\caption{\small  High-level planning evaluation for command following rate, within road rate and user preference rate.}
\vspace{-2mm}
\label{language_control}
\end{table}
\subsection{Implementation Details and Dataset}
\quad For the PedExecutor, we utilize Isaac Gym~\cite{makoviychuk2021isaac} as the physics simulation engine, with the AMASS dataset~\cite{mahmood2019amass} serving as the reference data clip for the discrimination reward. A model of the SMPL~\cite{loper2023smpl} robot as the simulation object. For the VehExecutor, we construct a simulation environment incorporating physical transitions. All LLM components used in the LLM-agents are implemented with GPT-4~\cite{achiam2023gpt} API. For the final rendering, we followed the rendering pipeline in ChatSim~\cite{wei2024editable}, using Blender for rendering with SMPL based human rendering implementation. Detailed experimental settings, configurations and video results are provided in the supplementary materials.

\subsection{System Results}

\quad We present keyframes from the output of the system in two scenes, as shown in Figure \ref{fig:system_result}. Each generated output includes abundant traffic participants based on user command. From the system's outputs, we see that: (i) interaction is thoroughly depicted, including human-vehicle interactions (e.g. taking a taxi, avoidance, slowing down), vehicle-vehicle interactions (e.g. lane changes, overtaking), and human-human interactions (e.g. pushing, chasing, walking with arm around shoulder). These interactions among different types of traffic participants contribute to a more diverse and realistic scene dynamics; (ii) the system achieves precise and detailed control, with each scene description being complex, potentially including abstract semantics. Through the design of multi-LLM-agents role-playing, complex instructions are accurately analyzed and executed, and abstract semantic information is effectively broken down into executable commands for final generation; (iii) the generated results are highly realistic. The physics-based control policies for pedestrians and vehicles produce realistic, and intuitive dynamics, particularly highlighting physical feedback in human-human interaction.
\begin{figure}[t]
    \centering
\includegraphics[width=0.45\textwidth]{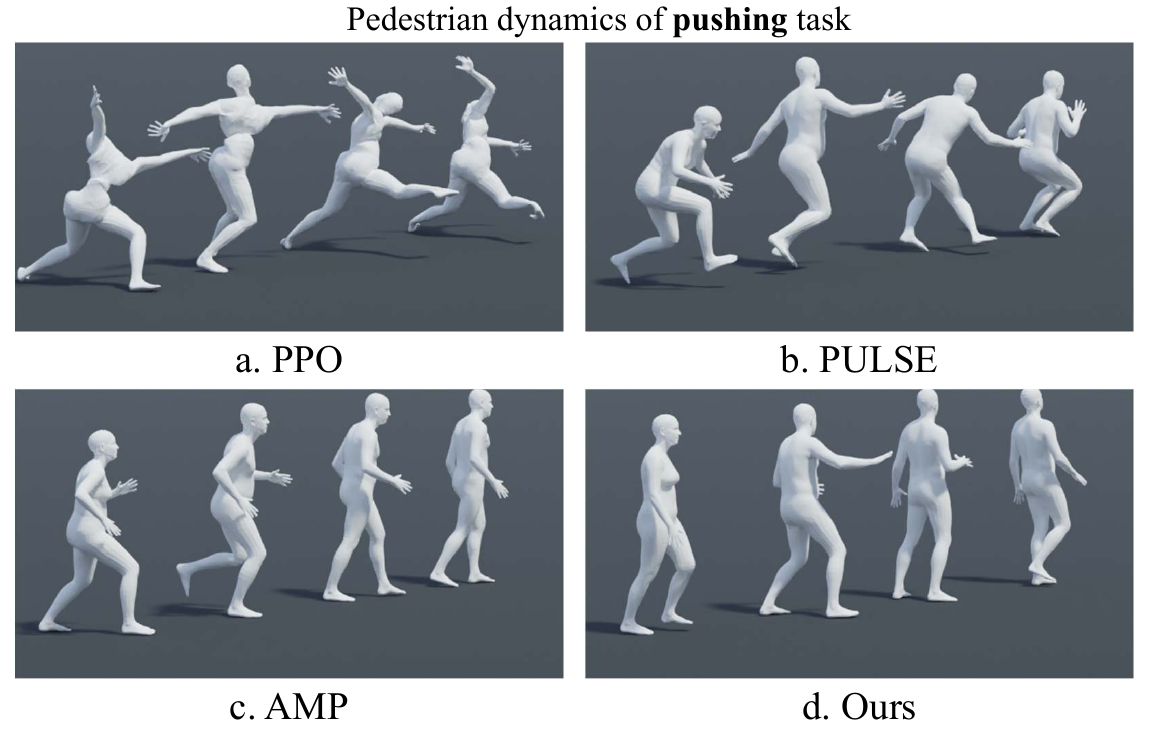}
\vspace{-3mm}
    \caption{\small Visualization comparison of interaction task execution.
    }
    \vspace{-1mm}
    \label{fig:pedestrian_compare}
\end{figure}

\begin{table}[t]
\centering
\footnotesize

\vspace{-2mm}
\resizebox{.99\columnwidth}{!}{
\begin{tabular}{c|ccccccc}
\toprule
Methods & FID$\downarrow$ & Div.$\uparrow$ & l-FID$\downarrow$ & l-Div.$\uparrow$ & $E_{mpjpe}$$\downarrow$ & $E_{f}$$\downarrow$ & UP$\uparrow$ \\
\midrule
Pacer\cite{rempe2023trace} & 7.25 & 1.24 & 7.93 & 1.05 & $\backslash$ & 0.122 & 0.179 \\
Pacer+\cite{wang2024pacer+} & 6.62 & 1.58 & 7.76 & 1.28 & 82.33 & 0.128 &  0.289 \\
Ours & \textbf{6.21} & \textbf{1.76} & \textbf{7.07} & \textbf{1.49} & \textbf{79.82} & 0.124 & \textbf{0.532}\\
\bottomrule
\end{tabular}
}
\caption{\small Dynamics quality, diversity and accuracy evaluation for trajectory following and motion imitation tasks.}
\vspace{-2mm}
\label{following_and_imitation}
\end{table}

\subsection{Component Results}

\subsubsection{LLM-agents planning}
\quad We compare the planned trajectory generated by LLM agents with existing traffic generation methods based on language, including LCTGen~\cite{tan2023language} and ChatSim~\cite{wei2024editable}. Since other methods do not support scenarios involving pedestrians, we conduct an evaluation under vehicle-only instructions. We define three types of instructions: single-vehicle instructions, interaction instructions, and composite instructions. Using 5 maps, we generate 26 samples per instruction type and invite 10 users to evaluate each generated result. They assess whether each result matches the language description, whether it is within-road, and their preference for each sample. The proportion of positive responses for each criterion is summarized in Table \ref{language_control}. The results indicate that ChatDyn consistently produced more accurate planning results that matched descriptions and were preferred by users. We also provide visual examples in Figure \ref{fig:planning_compare}, which also involves some planning results of ChatDyn with pedestrians. LCTGen struggles with interaction controls and complex instructions, while ChatDyn accurately fulfill all requirements, producing high-quality outputs that meet specifications, even in scenarios involving pedestrians.

\subsubsection{Pedestrian executor}

\quad We follow the evaluation from Pacer+~\cite{wang2024pacer+} to assess trajectory following and motion imitation tasks for PedExecutor as shown in Table \ref{following_and_imitation}. The quality and diversity of the generated results are measured by Frechet Inception Distance (FID)~\cite{heusel2017gans} and diversity metric (Div.)~\cite{wang2024pacer+} on normal speed, while l-FID and l-Div. are the metrics on low speed; imitation accuracy by Mean Per-Joint Position Error ($E_{mpjpe}$), and following accuracy by following error ($E_{f}$). Additionally, user preference (UP) is assessed by inviting 15 users to evaluate 33 segments of dynamics. With the support of hierarchical control and related training strategy, PedExecutor generates higher quality dynamics and demonstrated highly competitive performance in both following and imitation tasks.

\begin{table}[t]
\centering
\footnotesize

\vspace{-3mm}
\resizebox{.99\columnwidth}{!}{
\begin{tabular}{cc|ccc}
\toprule
Methods & Unified policy & Interaction 1 & Interaction 2 & Interaction 3 \\
\midrule
PPO~\cite{schulman2017proximal} & $\times$ & 0.971 & 0.934 & 0.914 \\
AMP~\cite{peng2021amp} & $\times$ & 0.234 & 0.179 & 0.108 \\
Pulse~\cite{luo2023universal} & $\times$ & 0.975 & 0.942 & 0.925 \\
Ours & $\checkmark$ & \textbf{0.982} & \textbf{0.977} & \textbf{0.971} \\
\bottomrule
\end{tabular}
}
\caption{\small Interaction tasks success rate evaluation. }
\label{ped_interact}
\end{table}

\begin{table}[t]
\centering
\footnotesize

\vspace{-3mm}
\resizebox{.82\columnwidth}{!}{
\begin{tabular}{ccccc}
\toprule
    TE   &   HC    & M. AMP &  Success rate  & User preference\\
    \midrule
    $\times$ & $\checkmark$  & $\checkmark$ & 0.403  & 0.120  \\
    $\checkmark$ & $\checkmark$  & $\times$ & 0.186  & 0.069  \\
    $\checkmark$ & $\times$  & $\checkmark$ & 0.948  & 0.344   \\
    $\checkmark$ & $\checkmark$ & $\checkmark$ & \textbf{0.977} & \textbf{0.467}   \\
    \bottomrule
\end{tabular}
}
\caption{\small PedExecutor interaction task ablation study. }
\vspace{-2mm}
\label{pedestrian_ablation}
\end{table}

We also evaluate the process of performing interaction tasks to determine each method's effectiveness in completing these tasks. Note that PedExecutor is a unified policy handling all tasks, while other methods rely on separately trained policies for each specific interaction task. Three types of interaction tasks are selected for training and validation: pushing (interaction 1), patting (interaction 2), and walking with an arm around the shoulder (interaction 3). The experimental results are presented in Table \ref{ped_interact}, with visual results shown in Figure \ref{fig:pedestrian_compare}.
From the both results,  we see that, AMP~\cite{peng2021amp} generates dynamics resembling the reference motion, but struggles to access suitable reference data clips for interaction tasks, resulting in actions that only superficially mimic the reference without fulfilling the task requirements. PPO~\cite{schulman2017proximal} and PULSE~\cite{luo2023universal} demonstrate relatively high task success rates, but the visualizations show that their outputs lack human-like style and appear unnatural. In contrast, with hierarchical control and  body masked AMP, PedExecutor achieves both high success rates and maintains high-quality, human-like outputs.

As shown in Table \ref{pedestrian_ablation}, we also conduct ablation studies on interaction tasks to verify the effectiveness of task embedding (TE), hierarchical control (HC), and body masked AMP (M. AMP). The results indicate that, without task embedding, PedExecutor struggles to accurately identify the required interaction task, leading to difficulty in successful task completion. When body masked AMP is removed, similar to AMP, the discrimination reward lacks a task-completing reference. In this situation, the outcomes can merely be close to the reference data clip without the ability to successfully execute the intended task. Without hierarchical control, the absence of action priors slows the training process and reduces the success rates. More video comparison results can be found in the supplementary materials.

\subsubsection{Vehicle executor}
\begin{figure}[t]
    \centering
\includegraphics[width=0.4\textwidth]{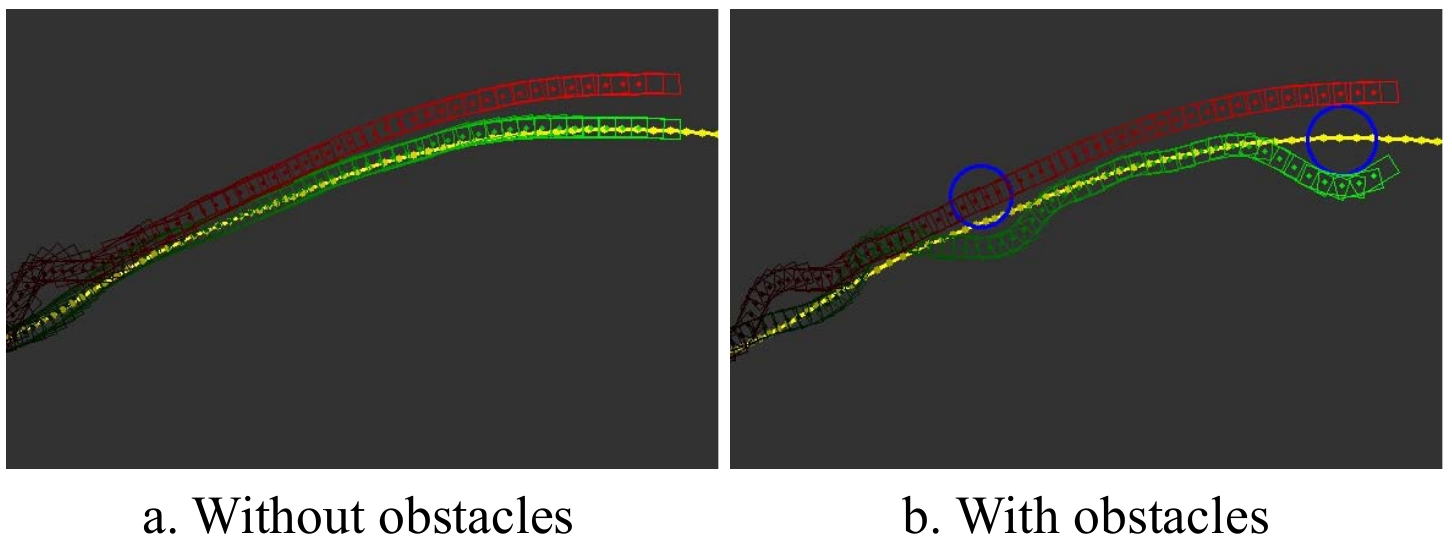}
\vspace{-3mm}
    \caption{\small Comparison of vehicle dynamics generation. \textcolor{green}{Green} boxes are our results and \textcolor{red}{red} boxes are Xu et al.'s~\cite{xu2023drl}. \textcolor{yellow}{Yellow} lines are the planned trajectory and \textcolor{blue}{blue} circles are obstacles. }
    \label{fig:vehicle_compare}
    \vspace{-2mm}
\end{figure}
\quad We evaluate the accuracy of VehExecutor under varying initial speeds by measuring the position error and velocity error, comparing our approach with Xu et al.'s method~\cite{xu2023drl} and pure pursuit~\cite{craig1992implementation} (PP) in Table \ref{veh_error}. VehExecutor consistently achieves the best performance across all speeds. We also conduct an ablation study in Table \ref{veh_ablation}, examining the effects of action filtering (Filt.), the composition of the action space (Action spa.) which means it consists directly of velocity and steering or their variations, and the inclusion of historical state (His. state) information. The ablation results validate the effectiveness of each design. Additionally, visual experiments in Figure \ref{fig:vehicle_compare} further demonstrate our method achieves more precise tracking, and obstacle avoidance capabilities in scenarios with obstacles.

\begin{table}[t]
\centering
\footnotesize

\vspace{-3mm}
\resizebox{.99\columnwidth}{!}{
\begin{tabular}{c|cccc}
    \toprule
    Methods/Speed     &   0    & 5  &  10 & 20 \\
    \midrule
    PP~\cite{craig1992implementation}        & 0.129/0.103 & 0.143/0.125 & 0.162/0.142 & 0.231/0.208 \\
    Xu et al.~\cite{xu2023drl} & 0.075/0.054 & 0.084/0.066 & 0.095/0.077 & 0.138/0.114 \\
    Ours      & \textbf{0.059}/\textbf{0.037} & \textbf{0.062}/\textbf{0.041} & \textbf{0.077}/\textbf{0.054} & \textbf{0.106}/\textbf{0.088}  \\
    \bottomrule
\end{tabular}

}
\caption{\small Position/velocity error of vehicle dynamics generation. }
\label{veh_error}
\end{table}

\begin{table}[t]
\centering
\footnotesize

\vspace{-3mm}
\resizebox{.85\columnwidth}{!}{
\begin{tabular}{ccccc}
    \toprule
    Filt.    &  Action spa.     & His. state & Position error & Velocity error\\
    \midrule
    $\times$ & $\times$  & $\times$ & 0.138  &  0.114  \\
    $\checkmark$ & $\times$  & $\times$ & 0.132  & 0.111      \\
    $\checkmark$ & $\checkmark$  & $\times$ & 0.115  & 0.092      \\
    $\checkmark$ & $\checkmark$  & $\checkmark$ & \textbf{0.106} & \textbf{0.088}      \\
    \bottomrule
\end{tabular}
}
\caption{\small VehExecutor ablation study. }
\label{veh_ablation}
\end{table}

\section{Conclusion and Limitations}
\quad We propose ChatDyn, the first system to achieve interactive and realistic language-driven multi-actor dynamics generation in street scenes. ChatDyn utilizes multi-LLM-agent role-playing to enable interaction-aware high-level planning under instructions. We introduce PedExecutor, a unified control policy for realistic pedestrian dynamics generation across multiple tasks, also enabling fine-grained interactions. We introduce VehExecutor, a physics-based vehicle control policy with a history-aware design to generate realistic vehicle dynamics.
The dynamics output can be used as a prior for various kinds of simulators including renderer, reconstruction, and generation model.

\textbf{In the future}, we plan to integrate more types of dynamics with diverse characteristics, such as cyclists and certain common animals, to reflect more realistic and varied scene dynamics. At the same time, more complex and diverse interaction behaviors can be further explored to construct scene information and events with greater accuracy. The relevant methodologies, after specific optimizations, can also be applied to more closed, yet finer-grained indoor scenes.

{
    \small
    \bibliographystyle{ieeenat_fullname}
    \bibliography{main}
}

\clearpage
\setcounter{page}{1}
\maketitlesupplementary
\appendix
\renewcommand{\thesection}{S\arabic{section}}
\renewcommand{\thefigure}{S\arabic{figure}}
\renewcommand{\thetable}{S\arabic{table}}
\setcounter{section}{0}
\setcounter{figure}{0}
\setcounter{table}{0}

\section{Details of PedExecutor}
\subsection{PedExecutor State Details}
The components of the humanoid's proprioception $S^{p}_{prop}$ are as follows: joint positions \( \mathbf{j} \in \mathbb{R}^{24 \times 3} \), rotations \( \mathbf{q} \in \mathbb{R}^{24 \times 6} \), linear velocities \( \mathbf{v} \in \mathbb{R}^{24 \times 3} \), and angular velocities \( \mathbf{\omega} \in \mathbb{R}^{24 \times 3} \)~\cite{wang2024pacer+}. These components are normalized with respect to the agent’s heading and root position in our simulator. The rotation \( \mathbf{q} \) is represented using a 6-degree-of-freedom rotation representation. $S^{p}_{prop}$ along with the trajectory state \( S^{p}_{traj} \), the motion state \( S^{p}_{mo} \) to be mimicked, and the target state \( S^{p}_{tar} \) for interaction, collectively forms the complete observation. During the task masking process, the remaining relevant states are masked by multiplying them by 0 based on the task to be executed, and in the motion state, specific joints can also be masked by multiplying them by 0 if necessary. For example, in the experiment, only the upper-body related states were retained.

\subsection{Task-related Reward and Training Details}
\textbf{Reward designs. }For trajectory following tasks~\cite{rempe2023trace}, the reward is defined as \( R_{\text{trajectory}} = e^{-2 \| \hat{p}_t - p_t \|} \), where $\hat{p}_t$ is the position to followed at $t$, $p_t$ is the current character position. For motion imitation tasks~\cite{luo2023perpetual}, \( R_{\text{imitation}} = e^{-100||\hat{j}_{pos}^t-j_{pos}^t||\odot m^t} + e^{-10||\hat{j}_{rot}^t-j_{rot}^t||\odot m^t} + e^{-0.1||\hat{j}_{vel}^t-j_{vel}^t||\odot m^t} + e^{-0.1||\hat{j}_{\omega}^t-j_{\omega}^t||\odot m^t}\), where $\hat{}$ indicates the motion states to be imitated, $m^t$ is the mask to select the joints for imitation, which are the joints of upper body in our experiments. Different interaction tasks require distinct reward designs~\cite{peng2022ase}; here, we present three sample tasks: for pushing, \( R_{\text{pushing}} = 1-u^{up} \cdot u\), $u^{up}$ is the global up vector u* is the target's up vector; for patting, \( R_{\text{patting}} = e^{-||p^{rh} - c||}\), $p^{rh}$ is the position of right hand and $c$ is the target contact position; and for walking while bending the shoulder, \( R_{\text{walking\_bending}} = e^{-2 \| \hat{p}_t - p_t \|} + e^{-||p^{rh} - c||} \), which combines the trajectory following and target position contacting. The final reward is calculated as \( R = 0.5 \cdot R_{\text{disc}} + 0.5 \cdot R_{\text{task}} \). 

\textbf{Interaction process. }The specific execution process for the three interaction tasks can be understood as follows: i) pushing, where the goal is to push the interaction object over, ii) patting, where the task is to gently tap a specific part of the interaction object (e.g., the shoulder) without knocking it over, and iii) walking with arm around another's shoulder, where the agent walks along a specific path while keeping the arm in contact with a specific location on the interaction object (e.g., the shoulder). During the training of the interaction tasks, the interaction object is replaced with a box in the physical environment to facilitate more stable training. However, during testing, the interaction object is replaced with another character in the simulation environment, and physical collisions and interactions are present, leading to the final output in the test phase.

\textbf{Failure recovery for robustness. }To enhance the robustness of PedExecutor, specifically to ensure that it can handle external disturbances without failing due to perturbations caused by physical collisions, we incorporate recovery during its training~\cite{luo2023perpetual}. In this approach, the human pose is initialized in a collapsed or otherwise unstable standing state. Training is then conducted from this initial state, allowing the policy to learn how to recover from failure. This enables the policy to exhibit robustness when responding to physical collisions and interactions. Without this, the policy might fail to continue the action after even minor disturbances.

\subsection{Network Architecture}
All relevant models, including the baselines, adopt the same network architecture, using an MLP as the policy network with hidden layers of 2048 and 1024 units. The final output is directed to either the latent space or the PD control signal, depending on whether hierarchical control is employed. The remaining network components, such as the discriminator, value network, control frequency (30Hz), and hyperparameters used for training, are consistent with those adopted in Pacer+~\cite{wang2024pacer+}. All training and testing are conducted on an NVIDIA 4090.

\subsection{Evaluation Details}
\textbf{Following and imitation. }For the following and imitation tasks, we adopt the same computation methods as those used in Pacer+~\cite{wang2024pacer+}. The calculations of FID and diversity are performed using the same manual feature extraction approach as in Pacer+, with 1000 segments selected from the AMASS dataset for FID computation. For the low-speed l-FID and l-diversity, we also follow Pacer+ by testing on instances where the speed is less than 1 m/s.

\textbf{Interaction tasks. }For the three interaction-related tasks: i) pushing, the success criterion is whether the object is pushed over within the specified timestep (with a tilt along the z-axis greater than 30°), and no part of the body other than the hands is in contact with the target; ii) patting, the success criterion is whether, within the specified timestep, the right hand is within 5 cm of the target's specific location and remains there for at least 50 timesteps, with no other part of the body in contact with the target except the hands; iii) walking with arm around another's shoulder, the success criterion is that the maximum deviation from the reference trajectory is no greater than 10 cm, and the right hand is within 5 cm of the target's specific location for at least 150 timesteps.
In the interaction ablation study, for the user study, we recruit 15 participants to test 30 dynamic sequences and select the ones they considered to have the highest quality.

\subsection{Rendering Details}
We utilize the rendering pipeline from ChatSim~\cite{wei2024editable}, employing Blender's Cycles as the rendering engine. Background rendering and HDRI lighting from ChatSim are applied to the scene. The human model is based on SMPL~\cite{loper2023smpl}, with the corresponding mesh initialized, and rendered using skin and clothing textures provided in the Bedlam dataset~\cite{black2023bedlam}.

\begin{figure}
    \centering
    \includegraphics[width=0.99\linewidth]{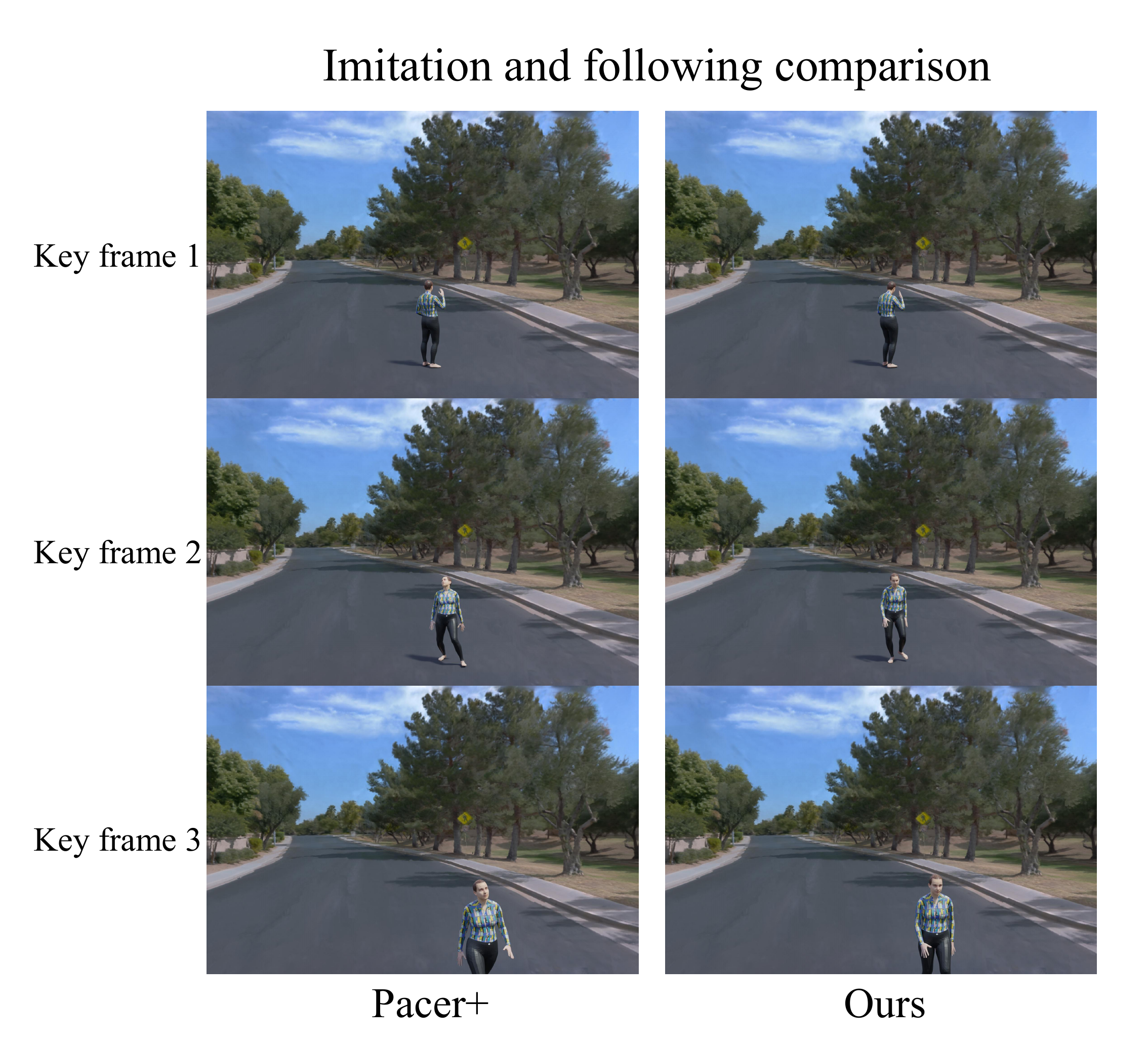}
    \caption{Comparison of imitation and following.}
    \label{im_fo_comp}
\end{figure}

\begin{figure}
    \centering
    \includegraphics[width=0.99\linewidth]{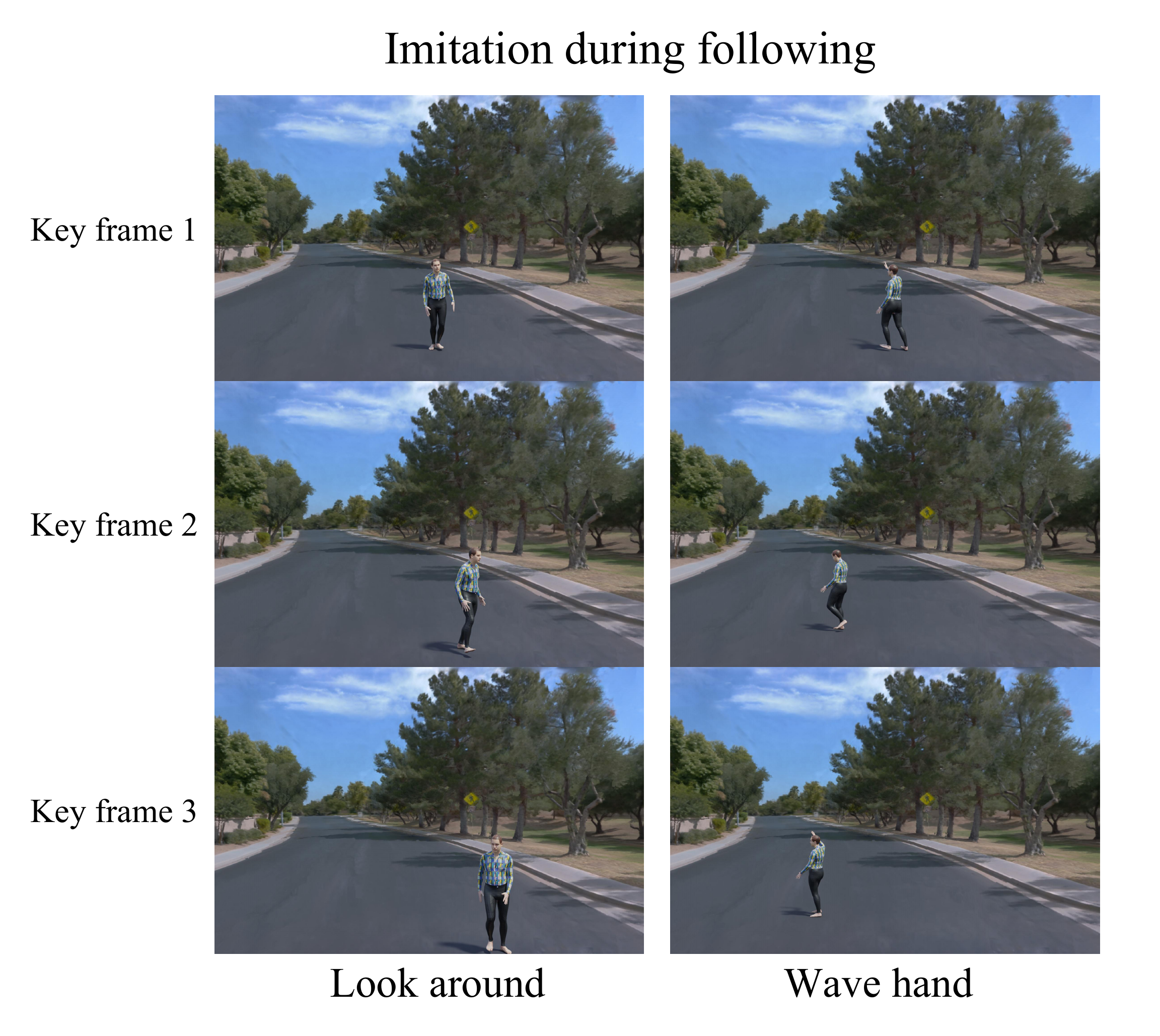}
    \caption{Imitation during following.}
    \label{im_during_fo}
\end{figure}

\begin{figure}
    \centering
    \includegraphics[width=0.99\linewidth]{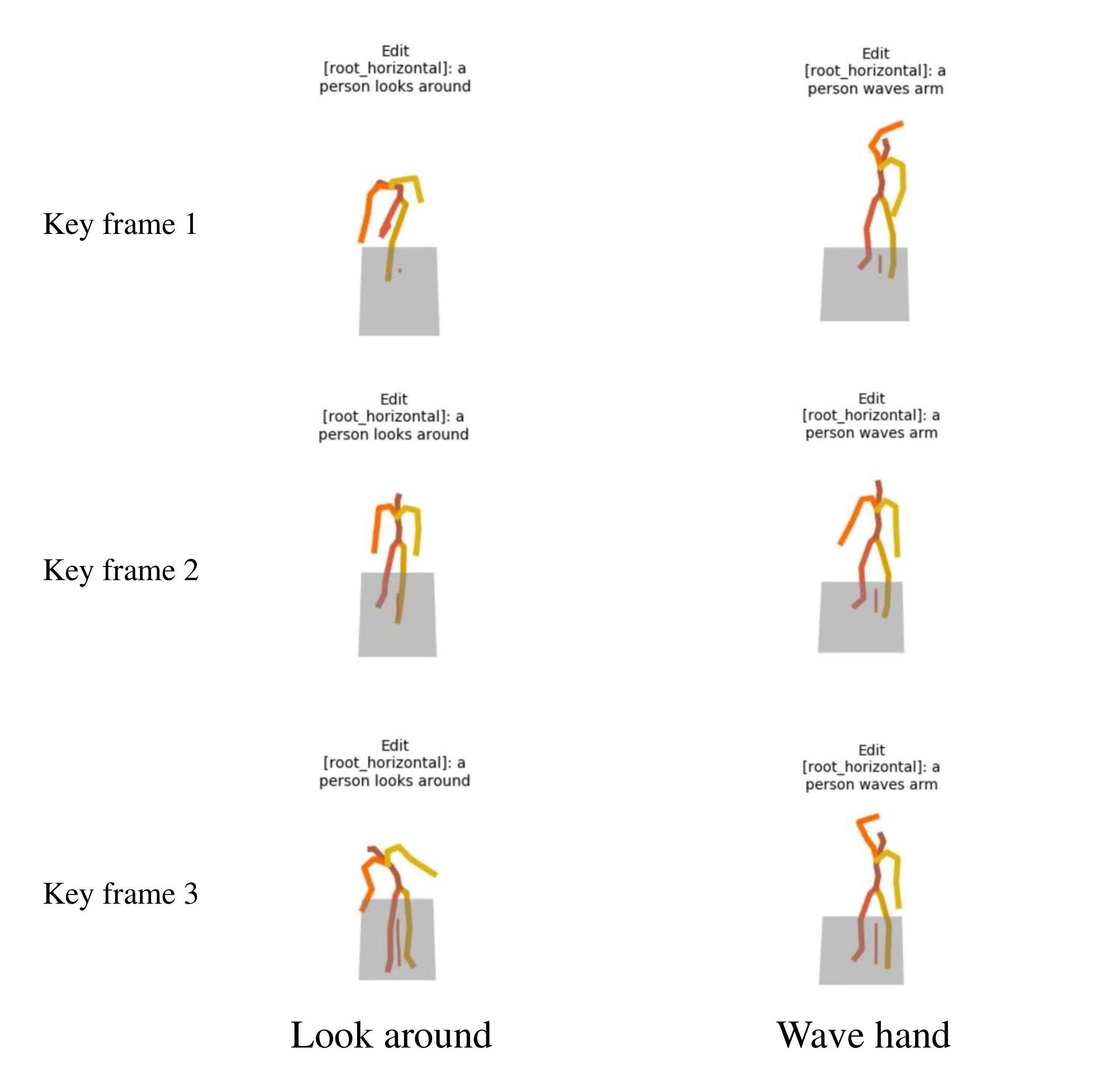}
    \caption{Failure of PriorMDM~\cite{shafir2023human} with action specification during following.}
    \label{priormdm}
\end{figure}

\subsection{Supplymentary Experiments}
We provide more comprehensive visualization results to compare different aspects of PedExecutor's characteristics. All comparisons, along with dynamic results, can also be observed in the supplementary video material.

\textbf{Imitation and following comparison. } As shown in \ref{im_fo_comp}, we compared the visual effects of Pacer+ during the imitation and following processes. It is evident that, under the influence of hierarchical control, PedExecutor enables smoother and more seamless transitions between different tasks—transitions that cannot be achieved with the priors provided by AMP.

\textbf{Results of imitation during following. }As shown in \ref{im_during_fo}, we also provide the results of PedExecutor performing both imitation and following simultaneously. In the case of following a specific trajectory, the upper body performs actions such as looking around and waving, showcasing capability of PedExecutor  to handle both imitation and following tasks concurrently.

\begin{figure*}
    \centering
    \includegraphics[width=0.99\linewidth]{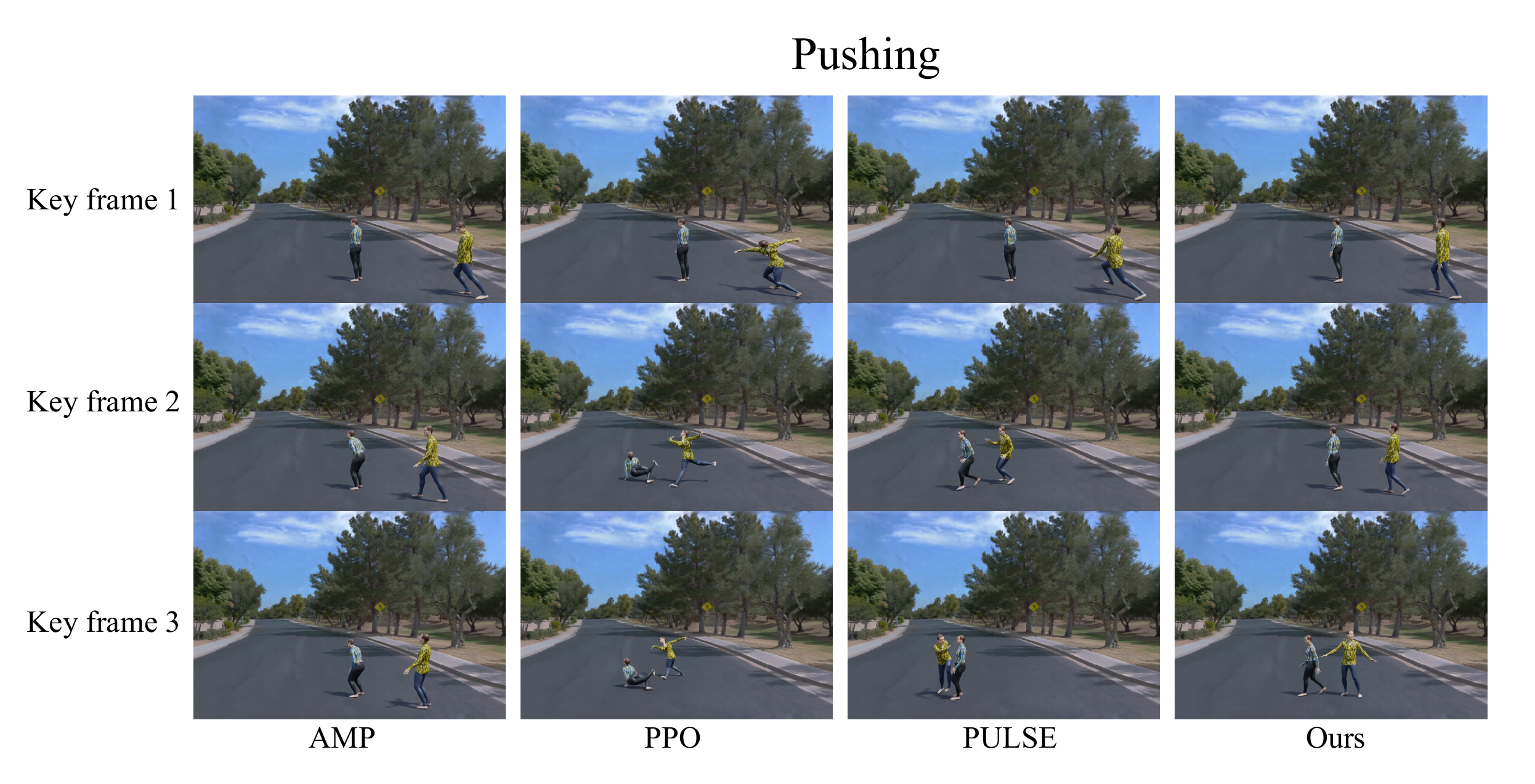}
    \caption{Comparison of pushing.}
    \label{pushing_comp}
\end{figure*}

\begin{figure*}
    \centering
    \includegraphics[width=0.99\linewidth]{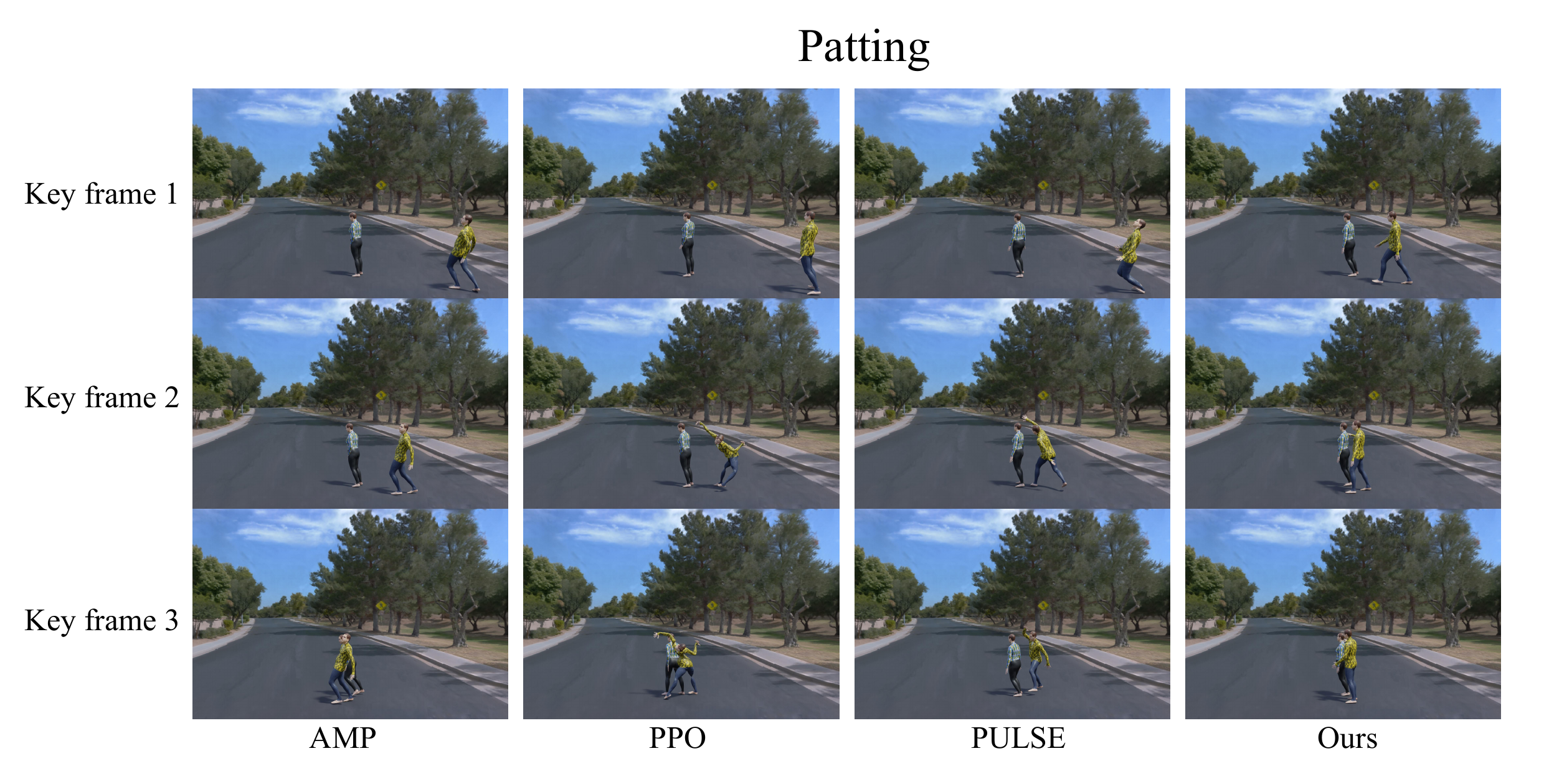}
    \caption{Comparison of patting.}
    \label{patting_comp}
\end{figure*}

\begin{figure*}
    \centering
    \includegraphics[width=0.99\linewidth]{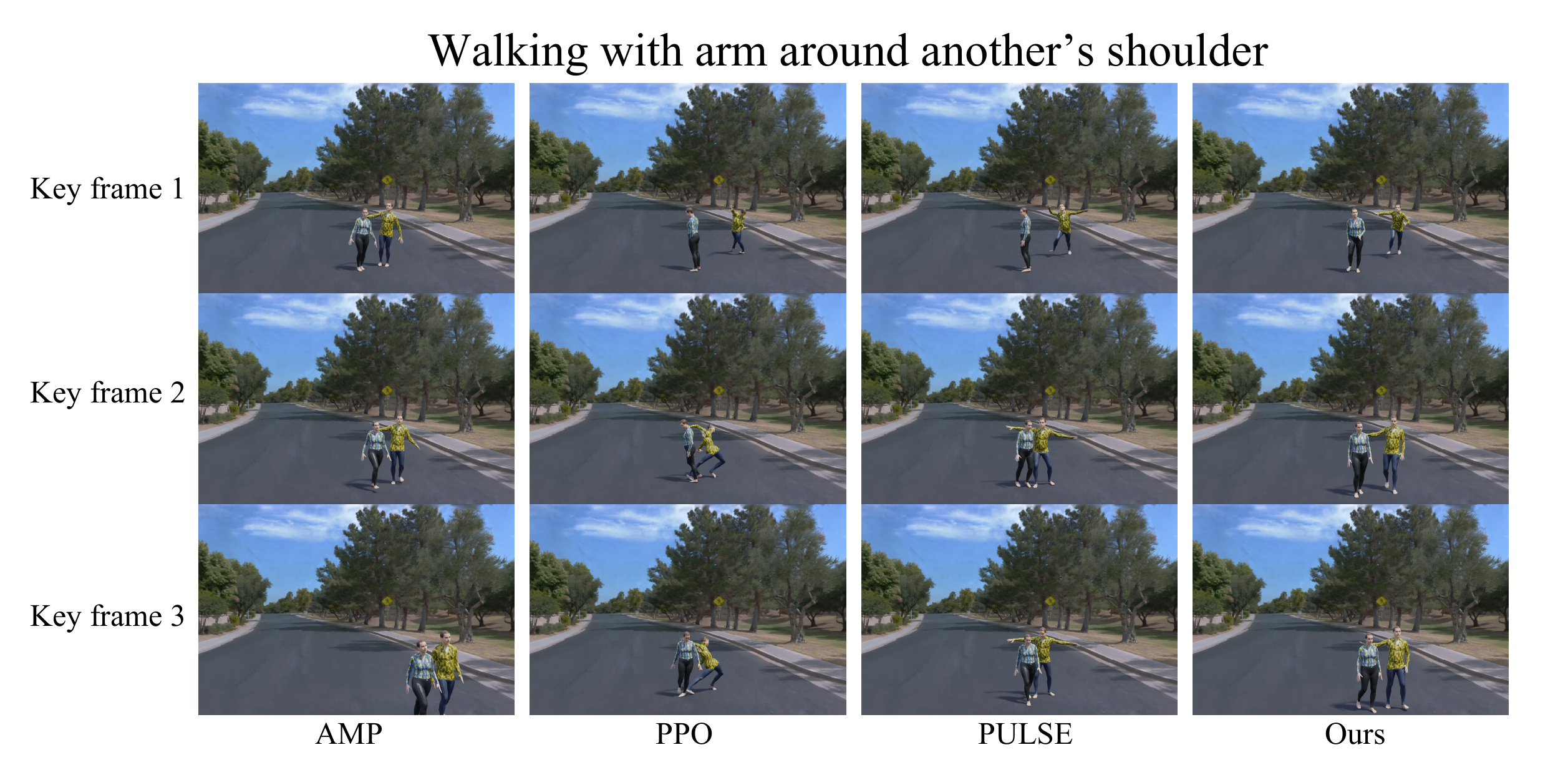}
    \caption{Comparison of walking with arm around another one's shoulder.}
    \label{walking_comp}
\end{figure*}

\textbf{Interaction task comparison. }As shown in \ref{pushing_comp} \ref{patting_comp} \ref{walking_comp}, we compared the performance of AMP, PPO, PULSE, and our method across three interaction tasks. It is still evident that while PPO and PULSE are able to complete the tasks under the given conditions, they fail to produce natural and human-like results. AMP, on the other hand, can only approximate the reference in the discriminator (walking or running), and is completely unable to complete the tasks. In contrast, our PedExecutor successfully completes the tasks while generating natural and human-like results.

\subsection{Further Discussion of Kinematics Methods}
For kinematics-based methods, some approaches can achieve following and motion specification, but they are completely unable to handle interaction-related tasks. Additionally, for following and motion specification tasks, these methods often suffer from overfitting to the dataset, leading to suboptimal performance. As shown in \ref{priormdm}, the results in noticeable sliding steps and unnatural movements.

\section{Details of VehExecutor}
\subsection{Network Architecture and Parameters of Bicycle-model}
All networks in VehExecutor are implemented as MLPs. The policy network consists of layers with dimensions 256, 256, 128, 128, 64, and 64, while the value network has layers with dimensions 1024, 512, 256, and 128. During training, the parameters of the bicycle model (L, W, \( l_f \), \( l_r \)) are set to two configurations: (2.7, 1.8, 0.9, 0.9) and (6.1, 2.5, 2.3, 2.0), mixed to accommodate vehicles of varying sizes. These parameters can be adjusted as needed based on specific requirements, with the two configurations provided here serving as examples.
\subsection{Obstacle state}
The states of obstacles are composed of their orientation and distance relative to the vehicle's own coordinate system (considering the radius of the obstacles). The state vector is initialized with a maximum number of observable obstacles. The vector is then populated with the specific identifiers of the actual obstacles, and any remaining entries are masked. Obstacles that are too far away are directly excluded, meaning their states are not considered, and they do not contribute to the reward calculation. The distance threshold for exclusion is set to 10 in the experiment.
\subsection{Supplymentary Experiments}
\textbf{Robustness. } We also provide the results using LQR~\cite{li2004iterative} in \ref{LQR}. It can be observed that LQR is capable of vehicle dynamics generation from a planned trajectory to some extent. However, the planned trajectory lacks relevant constraints, which may lead to unreasonable turns or abrupt changes, causing LQR to often produce less than ideal results. Furthermore, we tested the robustness of vehicle dynamics generation by adding noise with a mean of 0 and variance of \(\sigma\) to the planned trajectory. As shown in \ref{Robustness}, the results show that LQR is highly sensitive to noise, often producing significantly worse outcomes under its influence, while other methods are relatively less affected by the noise.
\begin{table}[t]
\centering
\footnotesize
\resizebox{.99\columnwidth}{!}{
\begin{tabular}{c|cccc}
    \toprule
    Methods/Speed     &   0    & 5  &  10 & 20 \\
    \midrule
    LQR~\cite{li2004iterative} & 0.074/0.058 & 0.086/0.070 & 0.092/0.079 & 0.125/0.0108 \\
    \bottomrule
\end{tabular}
}
\caption{\small Position/velocity error of LQR. }
\label{LQR}
\end{table}

\begin{table}[t]
\centering
\footnotesize

\resizebox{.99\columnwidth}{!}{
\begin{tabular}{c|cccc}
    \toprule
    $\sigma$    &  PP~\cite{craig1992implementation}    &  Xu et al.~\cite{xu2023drl}  &  LQR~\cite{li2004iterative}  &  ours \\
    \midrule
    0.00 & 0.162/0.142 & 0.095/0.077 & 0.092/0.079 & \textbf{0.077}/\textbf{0.054} \\
    0.01 & 0.168/0.147 & 0.098/0.080 & 0.322/0.289 & \textbf{0.079}/\textbf{0.058} \\
    0.03 & 0.169/0.151 & 0.098/0.079 & 0.568/0.479 & \textbf{0.082}/\textbf{0.060} \\
    \bottomrule
\end{tabular}
}
\caption{\small Vehicle dynamics generation under Gaussian noise. $\sigma$ indicates the variance of noise. }
\label{Robustness}
\end{table}

\textbf{Visualization for effectiveness. } In the supplementary video, we provide results comparing the planned trajectories without using VehExecutor and with VehExecutor. It is evident that, without the involvement of VehExecutor, the dynamics generated by simply calculating heading between consecutive frames of the directly planned trajectory are highly unnatural, exhibiting noticeable tail swings and abrupt changes. In contrast, the results using VehExecutor are much more realistic and natural. This demonstrates the necessity of VehExecutor, as trajectories without physical constraints are highly unnatural and impractical.

\section{Details of High-level Planning}
\subsection{LLM-agent details}
We provide the relevant sample prompts for the LLM agent in \ref{oracle_prompt} and \ref{actor_prompt}. All outputs from the LLM are in JSON format, and corresponding follow-up functions are used to convert the JSON outputs into the required data structures.

Similar to the rendering process, we also use the Waymo Open Dataset~\cite{sun2020scalability} as the planning dataset in all experiments, and the final results are presented based on this dataset.

\subsection{Different LLMs}
We further validated the impact of different LLMs on the results in \ref{different_llms}. We conducted experiments using GPT-3.5~\cite{ouyang2022training} and Llama-3~\cite{dubey2024llama} 70B (smaller models struggle to accurately execute the instructions). All other experimental settings remained consistent with those in the main text. It is evident that while other LLMs can handle the task to some extent, GPT-4~\cite{achiam2023gpt} demonstrates the most accurate understanding and decomposition of the instructions.
\begin{table}[t]
\centering
\footnotesize
\resizebox{.99\columnwidth}{!}{
\begin{tabular}{c|cccc}
\toprule
\multirow{2}*{Methods}  &   \multicolumn{3}{c}{Language command category} & \multirow{2}*{{Within road}}\\
~ & single & interaction & compound & ~   \\
\midrule
Ours-Llama3& 0.885 & 0.742 & 0.812 & 0.920  \\
Ours-GPT3.5 & 0.854 & 0.738 & 0.834 & 0.915\\
Ours-GPT4 & \textbf{0.952} & \textbf{0.883} & \textbf{0.896} & \textbf{0.935} \\
\bottomrule
\end{tabular}
}
\caption{\small  High-level planning evaluation for different LLMs.}

\label{different_llms}
\end{table}

\subsection{Collision handling}
In the high-level planning process, we also designed a collision handling mechanism to avoid unintended collisions. Specifically, during the trajectory generation, collision detection is performed, and when a collision is detected, a velocity adjustment function is applied to modify the speed of one of the agents. This velocity adjustment function uses a nonlinear mapping to combine the original planned result with an interpolated trajectory, leading to a planning result with different speeds. We evaluated the probability of collisions across 50 generated samples, which is calculated by dividing the number of vehicles that experienced a collision by the total number of vehicles. As shown in \ref{Collision}, the collision rate for all methods remains low, and ChatDyn also achieves a low collision rate while incorporating collision handling.
\begin{table}[t]
\centering
\footnotesize
\resizebox{.7\columnwidth}{!}{
\begin{tabular}{c|ccc}
    \toprule
    ~ & ChatSim~\cite{wei2024editable} & LCTGen~\cite{tan2023language} & Ours \\
    \midrule
    Collision rate & 0.149 & 0.092 & 0.067 \\
    \bottomrule
\end{tabular}
}
\caption{\small Collision rate of high-level planning }
\label{Collision}
\end{table}

\begin{figure*}

\begin{tcolorbox}[title = {Oracle agent prompt.}]
I have a requirement for analyzing a scenario. I will provide you with a requirement, and I need your help to break it down into four pieces of information: (1) identify all the agents, (2) initialize each agent's state, (3) formulate a text for each agent. You should provide the information in JSON format. Note that your output should only include the JSON format of the information, not the analysis process.

(1) Identify all the agents: This means you need to extract all the agents from the scenario based on the input text. The text may be quite lengthy, so you can locate the agents by identifying the nouns, which may assist you in this task.
The agents must be the objects involved in autonomous driving. The format should be as follows: agent\_list = \{'0': 'agent\_name\_0', '1': 'agent\_name\_1', ...\}. The keys (e.g., '0', '1', etc.) represent unique agent IDs, which you should assign starting from '0'. The agents should only include vehicles like cars, pedestrians, trucks, and buses, and should not include static objects like trees or buildings. Ensure that each agent is given a distinct name. 
For example, in the sentence 'car a wants to overtake car b', the nouns are 'car a' and 'car b'. Both are objects in autonomous driving scenarios. Therefore, the agent\_list should be formatted as follows: agent\_list = \{'0': 'car\_a', '1': 'car\_b'\}. Pedestrians can be represented with identifiers like `ped\_0` and may share numbering with entities such as `car\_0`.

(2) Initialize each agent's state. In this task, you need to determine four aspects for each agent: 1. Agent type, 2. Movement, \
            3. Speed. You can use the results from task (1) to complete this task. Provide the initial state of each agent in a list, formatted in JSON.\
            Most time the speed is bigger than 0.
For example, the initial states should be defined as follows: init-states = [{'agent\_id': '0', 'agent\_type': 'car', 'movement': 'overtaking',\
            'speed': 60}, {'agent\_id': '1', 'agent\_type': 'car', 'movement': 'straight', 'speed': 30}]. \
            If one car intends to overtake another, it should ideally be at least twice as fast as the car it is overtaking. 
            The initial state must include the agent type, movement,and speed.

Agent type should in [pedestrian,vehicle], action should in ['static', 'straight','pull over', 'turn over','overtake','turn left','turn right','straight left','straight right']\
            if it's a vehicle, and in ['static','crossing','straight'] if it is a pedestrian.

(3) Formulate a text for each agent. As an omniscient observer, you should instruct each agent on their actions through a text. The text should contain two pieces of information:\
            1. The agent type, 2. The agent's intention. You need to provide this in a JSON format. guide\_texts = {'0': 'text1', '1': 'text2', '2': 'text3', ...} \
                where the key is the agent's ID and the value is the text. The agent's name should be consistent with those in the agent\_list.

Your answer should be in a JSON format,and must include the three information:agent\_list,init-state,guide\_texts.And init-state must include the agent type,movement and speed.

\end{tcolorbox}
\caption{Oracle agent prompt.}
\label{oracle_prompt}
\end{figure*}

\begin{figure*}

\begin{tcolorbox}[title = {Actor agent prompt.}]
Now,you are an agent in the autonomous driving scenario.I will give you a text,and agent\_list, describing who you are and what you need to do. Note that your output should not include your analysis process, only the JSON format of the information you provide.

I need you to analyze the text and give me the four information of the ego agent to describe what type of lane the agent should be in:  (1)depend (2) speed change (3) keypoints list (4) behavior. Note that you only need to give the information of the ego agent,not the other agents.

(1) depend. You need to determine the depend of the agent according to the agent's intention. And give it like [depend\_agent\_id,depend\_type].You can get the depend\_agent\_id  from the agent\_list. Depend\_type should in ['end','start','trajectory','None'].'end' represents the ego agent's end point is the depend agent's end point, 'start' represents the ego agent's start point is the depend agent's start point, 'trajectory' represents the ego agent's whole trajectory is the depend agent's start point. If it has no depend, you should give it [-1,'None'].

(2) speed change. You need to determine the speed change of the agent according to the agent's intention. If you want to speed up, the speed change should be 1. If you want to slow down, the speed change should be -1. If you want to keep the speed, the speed change should be 0. For example, If there is someone nearby, you should slow down. 

(3) keypoints list.  You need to determine the number of keypoints required for your current behavior, the confirmation method for each keypoint, and their respective parameters, then return a list containing this information. Each keypoint can be confirmed in one of three ways: (1) Map-based (type `0`), with parameters: lane position (`left`, `right`, or `front`), lane type (`centerline` or `boundary`), and driving direction (`turn left`, `turn right`, or `straight`). (2) Lane-relationship-based (type `1`), with the parameter being the relationship to the previous keypoint (`opposite direction adjacent`, `same direction adjacent`, `adjacent straight`, `adjacent left turn`, `adjacent right turn`, `different type adjacent`, or `opposite boundary`). (3) Agent-based (type `2`), with parameters specifying required information type (`point` or `trajectory`). The result should be a sequential list of dictionaries where the key is the type (`0`, `1`, or `2`) and the value is the corresponding parameters. For example, a left-turning car might return  `[{'0': {'position': 'front', 'lane\_type': 'centerline', 'direction': 'turn left'}}, {'1': {'relationship': 'adjacent straight'}}]`, while a car in the left lane picking up `ped\_a` might return `[{'0': {'position': 'left', 'lane\_type': 'centerline', 'direction': 'straight'}}, {'2': {'info\_type': 'point'}}]`. Ensure compliance with parameters, common sense, and traffic regulations, with the first keypoint typically using type `0`. 

(4) behavior. You need to determine your behavior. If you are a vehicle, the behavior is "None." If you are a pedestrian, the behavior corresponds to the description provided. There are two scenarios: if your behavior is a specific action such as calling or waving, simply return the text of that action; if the behavior involves interactive actions such as pushing, patting, or walking with an arm around another person, you must return the exact predefined descriptions for these three types of interactions without modification. For example, if you are calling, your behavior is "calling phone."; if you push ped\_1, your behavior is "pushing."

Your answer should be in a JSON format, and must include depend, speed change, keypoints list and behavior.
\end{tcolorbox}
\caption{Actor agent prompt.}

\label{actor_prompt}
\end{figure*}

\end{document}